\documentclass{article}

\PassOptionsToPackage{numbers, compress}{natbib}

    \usepackage[preprint]{neurips_2022}

\usepackage[utf8]{inputenc} 
\usepackage[T1]{fontenc}    
\usepackage[colorlinks=true,linkcolor=red]{hyperref}       
\usepackage{url}            
\usepackage{booktabs}       
\usepackage{amsfonts}       
\usepackage{nicefrac}       
\usepackage{microtype}      
\usepackage{xcolor}         
\usepackage{graphicx}
\usepackage{subfig}
\usepackage{array}
\usepackage{multirow}
\usepackage{makecell}
\usepackage{booktabs}
\usepackage{hyperref}
\usepackage{svg}
\usepackage{hhline}
\usepackage{tabularx}
\usepackage{bbold}
\usepackage{mathtools}
\usepackage{adjustbox}
\usepackage{color, colortbl}
\usepackage{blindtext}
\usepackage{floatrow}
\newfloatcommand{capbtabbox}{table}[][\FBwidth]

\title{EnergyMatch: Energy-based Pseudo-Labeling for Semi-Supervised Learning}

\author{%
  Zhuoran Yu \hspace{20pt} Yin Li  \hspace{20pt} Yong Jae Lee\\\\
  University of Wisconsin-Madison\\
  }

\begin{document}

\maketitle

\vspace{-10pt}
\begin{abstract}
\vspace{-5pt}
  Recent state-of-the-art methods in semi-supervised learning (SSL) combine consistency regularization with confidence-based pseudo-labeling. To obtain high-quality pseudo-labels, a high confidence threshold is typically adopted.  However, it has been shown that softmax-based confidence scores in deep networks can be arbitrarily high for samples far from the training data, and thus, the pseudo-labels for even high-confidence unlabeled samples may still be unreliable.  In this work, we present a new perspective of pseudo-labeling: instead of relying on model confidence, we instead measure whether an unlabeled sample is likely to be ``in-distribution''; i.e., close to the current training data. To classify whether an unlabeled sample is ``in-distribution'' or ``out-of-distribution'', we adopt the energy score from out-of-distribution detection literature. As training progresses and more unlabeled samples become in-distribution and contribute to training, the combined labeled and pseudo-labeled data can better approximate the true distribution to improve the model. Experiments demonstrate that our energy-based pseudo-labeling method, albeit conceptually simple, significantly outperforms confidence-based methods on imbalanced SSL benchmarks, and achieves competitive performance on class-balanced data. For example, it produces a 4-6\% absolute accuracy improvement on CIFAR10-LT when the imbalance ratio is higher than 50. When combined with state-of-the-art long-tailed SSL methods, further improvements are attained.
\end{abstract}


\vspace{-10pt}
\section{Introduction}
\vspace{-5pt}

In semi-supervised learning (SSL)~\citep{zhu2005semi, zhu2009introduction, chapelle2009semi}, a machine learning model is trained with (a small amount of) labeled data and (a large amount of) unlabeled data, with the goal of reducing the cost of human annotation. In recent years, the frontier of SSL has seen significant advances through pseudo-labeling~\citep{rosenberg2005semi,lee2013pseudo} combined with consistency regularization~\citep{laine2016temporal, tarvainen2017mean, berthelot2019mixmatch, berthelot2019remixmatch, sohn2020fixmatch, xie2020unsupervised}. 

\begin{figure}[t]
\vspace{-1cm}
    \centering
    \includegraphics[width=.98\linewidth]{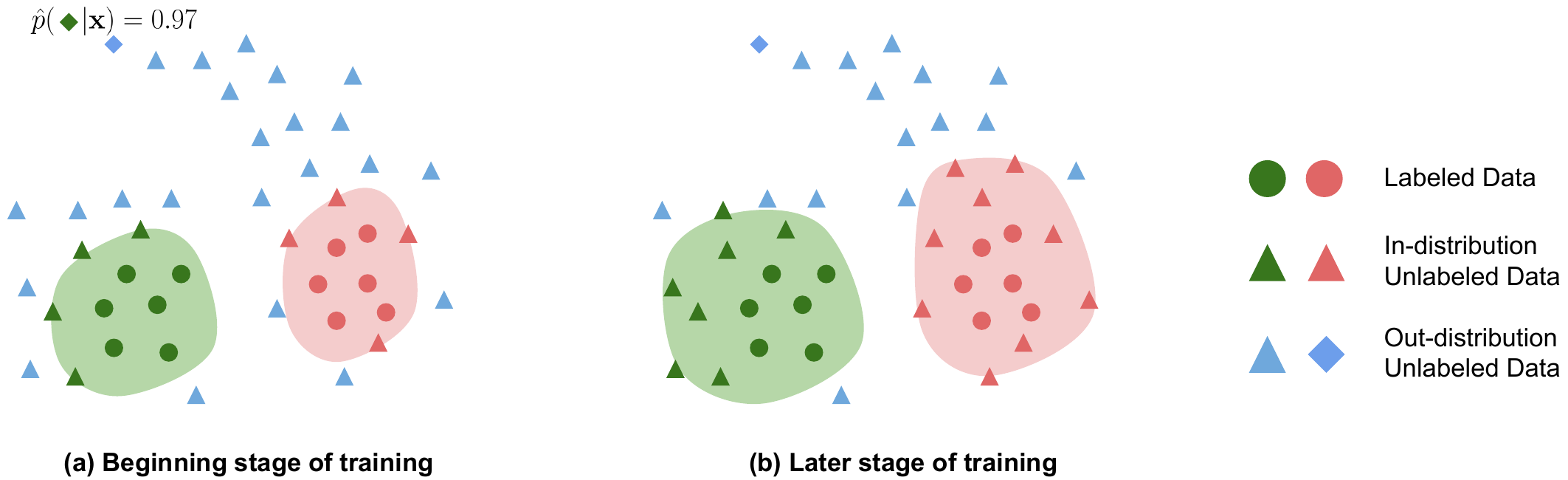}
    \vspace{-3pt}
    \caption{
    (a) At the beginning of training, only a few unlabeled samples are close enough to the training distribution formed by the initial labeled data. Note that with a confidence-based approach, the diamond unlabeled sample would be added as a pseudo-label for the green class since the model's confidence for it is very high (0.97).  Our energy-based method instead ignores it since its energy score is too high and is thus considered  out-of-distribution at this stage. (b) As training progresses, the training distribution is evolved by both the initial labeled data and the pseudo-labeled ``in-distribution'' unlabeled data, and more unlabeled data can be included in training.  In this toy example, with our energy-based approach, the diamond sample would eventually be pseudo-labeled as the red class.
    }
    \label{fig:illustration}
\end{figure}

Pseudo-labeling, a type of self-training~\citep{scudder1965probability,mclachlan1975iterative} technique, converts model predictions on unlabeled samples into soft or hard labels as optimization targets, while consistency regularization~\citep{laine2016temporal, tarvainen2017mean, berthelot2019mixmatch, berthelot2019remixmatch, sohn2020fixmatch, xie2020unsupervised} trains the model to produce the same pseudo-label for two different views (strong and weak augmentations) of an unlabeled sample.  State-of-the-art methods rely on confidence-based thresholding~\citep{lee2013pseudo, sohn2020fixmatch, xie2020unsupervised, zhang2021flexmatch} for pseudo-labeling, in which only the unlabeled samples whose predicted class confidence surpasses a very high threshold (e.g., 0.95) are pseudo-labeled for training. While this typically leads to high precision in the pseudo-labels, it can lead to low recall, especially in low-data settings (e.g., for rare classes in the imbalanced scenario~\citep{wei2021crest}). More critically, prior studies~\citep{szegedy2013intriguing,nguyen2015deep, hein2019relu} have shown that softmax-based confidence scores in deep networks can be arbitrarily high for samples that are far from the training data --- thus, in the SSL setting, the pseudo-label for any unlabeled sample that is far from the labeled training data may be unreliable, even if the model's confidence for it is very high. Figure\ \ref{fig:illustration} illustrates such an example.

In this work, we present a novel perspective for pseudo-labeling in SSL. Instead of relying on a model's prediction confidence to decide whether to pseudo-label an unlabeled instance or not, we propose to view the pseudo-labeling decision as an evolving in-distribution vs.\ out-of-distribution classification problem. Specifically, we treat instances that are close to the current training distribution---and hence likely to have more reliable predictions---as ``\emph{in-distribution}'', and those that are far---and hence likely to have unreliable predictions---as ``\emph{out-of-distribution}''.\footnote{Note that our definition of out-of-distribution is different from the typical definition from the Out-of-Distribution literature that constitutes unseen classes.}  At the beginning of training, only the unlabeled instances that are close to the initial labeled data will be treated as ``\emph{in-distribution}''.  As training progresses, more and more unlabeled instances are pseudo-labeled. The in-distribution vs.\ out-of-distribution boundary will evolve to be jointly shaped by both the initial labeled samples as well as the unlabeled samples that are pseudo-labeled up to that point.  Thus, unlabeled instances that were previously predicted to be out-of-distribution, could now be predicted as in-distribution as the distribution of the (pseudo-)labeled training data is continuously updated and expanded.


To determine whether an unlabeled sample is in-distribution or out-of-distribution, we leverage the energy score~\citep{lecun2006tutorial} for its simplicity and good empirical performance. The energy score is a non-probabilistic scalar that is derived from a model's output and theoretically aligned with the probability density of a data sample---lower/higher energy reflects data with higher/lower likelihood of occurrence following the training distribution, and has been shown to be useful for conventional out-of-distribution (OOD) detection~\citep{liu2020energy}.  In our SSL setting, at each training iteration, we compute the energy score for each unlabeled sample and pseudo-label it if its energy is below a certain threshold. If it is, then we choose its pseudo-label to be the predicted class made by the model. 

%
To the best of our knowledge, our work is the first to consider pseudo-labeling in SSL from an \emph{in-distribution vs.\ out-distribution} perspective. Our energy-based pseudo-labeling can easily replace vanilla confidence-based pseudo-labeling in any SSL framework. When integrated into FixMatch~\citep{sohn2020fixmatch}, our method performs significantly better than standard confidence-based pseudo-labeling methods when the training data is imbalanced across categories, which we believe better reflects real-world data distributions. For example, our method outperforms state-of-the-art methods by 4-6\% absolute accuracy on long-tailed CIFAR10~\citep{krizhevsky2009learning} when the imbalance ratio between the head class and tail class is greater than 50. 
When combining our energy-based pseudo-labeling with ABC~\citep{lee2021abc}, a method designed specifically for the long-tailed SSL setting, our approach still shows noticeable improvement, which demonstrates its generalizability to different SSL frameworks. Finally, on standard SSL benchmarks (CIFAR10/CIFAR100~\citep{krizhevsky2009learning}, STL-10~\citep{coates2011analysis}, and SVHN~\citep{netzer2011reading}) where training data is mostly balanced across the categories, our method shows either comparable or better performance compared to highly-competitive baselines.  This is in contrast to the existing SSL literature, in which an SSL method either performs well in the balanced or imbalanced setting, but not both.


\vspace{-5pt}
\section{Related Work}
\vspace{-5pt}

\textbf{Semi-Supervised Learning.} Research in SSL emerged a few decades ago~\citep{scudder1965probability, mclachlan1975iterative}.  Since then various directions have been proposed including entropy minimization~\citep{grandvalet2004semi}, graph-based methods~\citep{zhu2002learning, iscen2019label, song2022graph}, and virtual adversarial training~\citep{miyato2018virtual}. Recent advances in self-training~\citep{lee2013pseudo, xie2020self} and consistency regularization~\cite{laine2016temporal, tarvainen2017mean} have significantly pushed its frontier. In particular, MeanTeacher~\citep{tarvainen2017mean} proposes a teacher-student framework where the teacher is updated with an exponential moving average of a student model. Mixup-based methods~\citep{berthelot2019mixmatch, berthelot2019remixmatch} incorporate Mixup~\citep{zhang2017mixup} into consistency regularization. FixMatch~\citep{sohn2020fixmatch} and UDA~\citep{xie2020unsupervised} predict pseudo-labels on weakly-augmented views of unlabeled images and train the model to predict those pseudo labels on strongly-augmented views. Most of these state-of-the-art methods use confidence thresholding to retain high-quality pseudo-labels~\citep{sohn2020fixmatch, xie2020unsupervised}. Few works have attempted to revise the design of the pseudo-labeling approach. FlexMatch~\citep{zhang2021flexmatch} uses curriculum learning to dynamically adjust the confidence thresholds for different classes, while UPS~\citep{rizve2021defense} uses an uncertainty estimation with MC-Dropout~\citep{gal2016dropout}, in addition to the confidence score to select pseudo-labels. Our work proposes a different approach for pseudo-labeling. While prior works consider the confidence or uncertainty of model predictions, ours is the first to view the pseudo-labeling process from an ``in-distribution vs.\ out-of-distribution'' perspective. In particular, unlike UPS~\citep{rizve2021defense}, which uses the disagreement between a model ensemble to measure uncertainty in the pseudo-label for an unlabeled instance, we instead use the energy score from a single model's output to estimate its likelihood of occurrence. We show that this leads to better performance for SSL while being more computationally efficient. 

\textbf{Class-Imbalanced Semi-Supervised Learning.} While SSL has been extensively studied in the balanced setting in which all categories have (roughly) the same number of instances, class-imbalanced SSL has only begun to be explored recently. A key challenge is to avoid overfitting to the majority classes while capturing the minority classes. Prior works have devised different approaches tailored for this setting.
DARP~\citep{kim2020distribution} refines the pseudo-labels through convex optimization targeted specifically for the imbalanced scenario. CReST~\citep{wei2021crest} achieves class-rebalancing by pseudo-labeling unlabeled samples with frequency that is inversely proportional to the class frequency. ABC~\citep{lee2021abc} introduces an auxiliary classifier that is trained with class-balanced sampling. Our method is in parallel to these developments; it can be easily integrated into those prior approaches by replacing the confidence-based thresholding with an energy-based one, and achieve significant performance gain. 


\textbf{Out-of-Distribution Detection.} OOD detection aims to detect outliers that are substantially different from the training data, and is important when deploying ML models to ensure safety and reliability in real-world settings. The softmax score was used as a baseline for OOD detection in~\citep{hendrycks2016baseline} but has since been proven to be an unreliable measure by subsequent work~\citep{hein2019relu, liu2020energy}. Improvements have been made in OOD detection through temperatured softmax~\citep{liang2017enhancing} and the energy score~\citep{liu2020energy}. Exploring new methods in OOD detection is not the focus of our work. Instead, we show that leveraging the concept of OOD detection, and in particular, the energy score~\cite{lecun2006tutorial,liu2020energy} for pseudo-labeling, provides a new perspective in SSL that results in competitive and robust performance.
\vspace{-5pt}
\section{Approach}
\vspace{-5pt}

Our goal is to devise a more reliable way of pseudo-labeling in SSL, which accounts for whether an unlabeled data sample can be considered ``in-distribution'' or ``out-of-distribution'' based on the existing set of (pseudo-)labeled samples. We first overview the framework for state-of-the-art SSL methods that combine consistency regularization with confidence-based pseudo-labeling~\cite{sohn2020fixmatch, zhang2021flexmatch, xie2020unsupervised},
as our proposed approach simply replaces one step --- the pseudo-labeling criterion.  We then provide details on our energy-based solution.

\vspace{-2pt}
\subsection{Background: Consistency Regularization with Confidence-based Pseudo-Labeling}
\vspace{-2pt}

The training of pseudo-labeling SSL methods for image classification involves two loss terms: the supervised loss $\mathcal{L}_s$ computed on human-labeled data and the unsupervised loss $\mathcal{L}_u$ computed on unlabeled data. The supervised loss is typically the standard multi-class cross-entropy loss computed on weakly-augmented views (e.g., flip and crop)
of labeled images. Let $\mathcal{X} = \{(\mathbf{x_b}, \mathbf{y_b})\}_{b=1}^{B_s}$ be the labeled set where $\mathbf{x}$ and $\mathbf{y}$ denote the data sample and its corresponding one-hot label, respectively. Denote $p(\mathbf{y}|\omega(\mathbf{x_b}))=f(\omega(\mathbf{x_b}))$ as the predicted class distribution on input $\mathbf{x_b}$, where $\omega$ is a weakly-augmenting transformation and $f$ is a classifier often realized as a deep network. Then at each iteration, the supervised loss for a batch $B_s$ of labeled data is given by
\begin{equation}
    \mathcal{L}_s = \frac{1}{B_s}\sum_{b=1}^{B_s} \mathcal{H}(\mathbf{y_b}, p(\mathbf{y}|\omega(\mathbf{x_b}))),
\end{equation}
where $\mathcal{H}$ is the cross-entropy loss.

Mainstream research in SSL focuses on how to construct the unsupervised loss. One dominating approach is consistency regularization~\cite{laine2016temporal, tarvainen2017mean, berthelot2019mixmatch, berthelot2019remixmatch}, which regularizes the network to be less sensitive to input or model perturbations by enforcing consistent predictions across different views (augmentations) of the same training input, through an MSE or KL-divergence loss. Self-training~\citep{lee2013pseudo, rosenberg2005semi, rizve2021defense, xie2020self} converts model predictions to optimization targets for unlabeled images. In particular, pseudo-labeling~\citep{lee2013pseudo} converts model predictions into hard pseudo-labels. To ensure high quality of pseudo-labels, a high confidence threshold is often used. 

The recently introduced weak-strong data augmentation paradigm~\citep{sohn2020fixmatch, xie2020unsupervised} can be viewed as the combination of these two directions. When combined with confidence-based pseudo-labeling~\citep{sohn2020fixmatch, zhang2021flexmatch}, at each iteration, the process can be summarized as follows: 
\begin{enumerate}
    \item For each unlabeled data point $\mathbf{x}$, the model makes prediction $p(\mathbf{y}|\omega(\mathbf{x}))=f(\omega(\mathbf{x_b}))$ on its weakly-augmented view $\omega(\mathbf{x})$.
    \item Confidence thresholding is applied and a pseudo-label is only produced when the maximum predicted probability $\max_i p(y_i|\omega(\mathbf{x}))$ of $\mathbf{x}$ is above a threshold $\tau_c$ (typically, $\tau_c=0.95$).
    \item The model is then trained with its strongly-augmented view $\Omega(\mathbf{x})$ (e.g., RandAugment~\citep{cubuk2020randaugment} and CutOut~\citep{devries2017improved}) 
    along with its one-hot thresholded pseudo-label $\hat{p}(\mathbf{y}|\omega(\mathbf{x}))$  obtained on the weakly-augmented view.
\end{enumerate}
With batch size $B_u$ for unlabeled data, the unsupervised loss is formulated as follows:
\begin{equation}
    \mathcal{L}_u = \frac{1}{B_u}\sum_{b=1}^{B_u} \mathbb{1} [\max_i(p(y_i|\omega(\mathbf{x_b}))) \geq \tau_c] ~\mathcal{H}(\hat{p}(\mathbf{y}|\omega(\mathbf{x_b})), p(\mathbf{y}|\Omega(\mathbf{x_b}))),
\end{equation}
where $\mathbb{1}[\cdot]$ is the indicator function.

The final loss at each training iteration is computed by $\mathcal{L} = \mathcal{L}_s + \lambda \mathcal{L}_u$ with $\lambda$ as a hyperparameter to balance the loss terms. The model parameters are updated with this loss after each iteration. Our proposed energy-based pseudo-labeling can be integrated into the above process by replacing the confidence thresholding in step 2, as we will introduce in the next section.

\begin{figure}[t]
\vspace{-1cm}
    \centering
    \includegraphics[width=\linewidth]{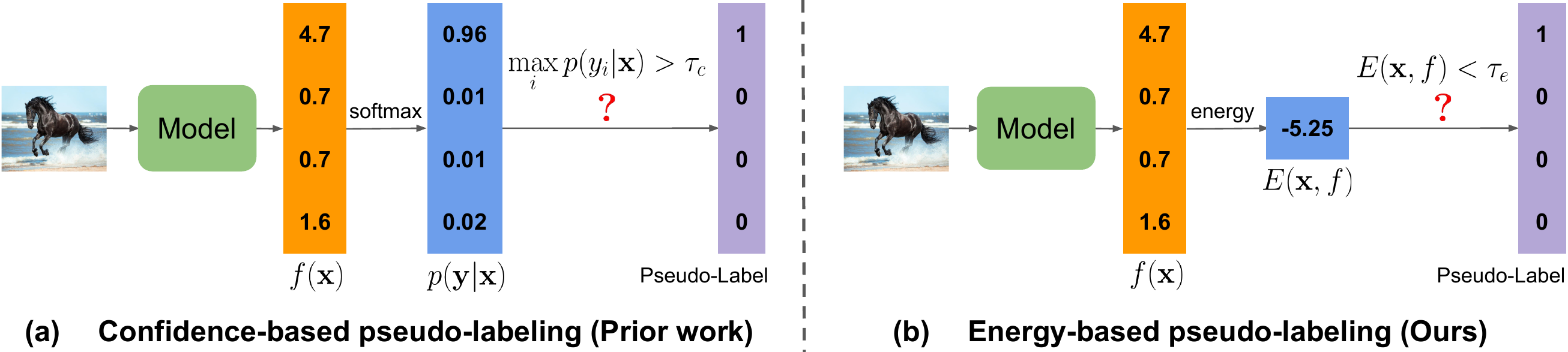}
    \caption{Overview of confidence-based pseudo-labeling vs.\ energy-based pseudo-labeling.}
    \label{fig:overview}

\end{figure}

\vspace{-2pt}
\subsection{Consistency Regularization with Energy-based Pseudo-Labeling}
\vspace{-2pt}

Although confidence-based thresholding typically leads to high precision pseudo-labels, it often leads to low recall in low-data settings (e.g., in the long-tailed scenario~\citep{wei2021crest} for tail classes). More critically, softmax-based confidence scores in deep networks are oftentimes overconfident~\citep{guo2017calibration}, and can be arbitrarily high for samples that are far from the training data~\citep{nguyen2015deep,hein2019relu}. The implication in the SSL setting is that the pseudo-label for even high-confidence unlabeled samples may not be trustworthy if those samples are far from the labeled data. 


To address this issue, our method tackles the pseudo-labeling process from a different perspective: instead of generating pseudo-labels for high-confidence samples, we produce pseudo-labels only for unlabeled samples that are close to the current training distribution --- we call these ``in-distribution'' samples. The rest are ``out-of-distribution'' samples, for which the model's confidences are deemed unreliable.  The idea is that, as training progresses and more unlabeled samples become in-distribution and contribute to training, the training distribution will better approximate the true distribution to improve the model, and in turn, improve the overall reliability of the pseudo-labels (see Figure~\ref{fig:illustration}). 

To determine whether an unlabeled sample is in-distribution or out-of-distribution, we use the energy score~\citep{lecun2006tutorial} derived from the classifier $f$. The energy score is defined as:
\begin{equation}
\label{eq:energy}
    E(\mathbf{x}, f(\mathbf{x})) = -T \cdot \log(\sum_{i=1}^K e^{f_i(\mathbf{x} / T)}),
\end{equation}
where $\mathbf{x}$ is the input data and $f_i(\mathbf{x})$ indicates the corresponding logit value of the $i$-th class. $K$ is the total number of classes and $T$ is a tunable temperature. 

When used for conventional OOD detection, smaller/higher energy scores indicate that the input is likely to be in-distribution/out-of-distribution. Indeed, a discriminative classifier implicitly defines a free energy function~\citep{grathwohl2019your} that can be used to characterize the data distribution~\citep{lecun2006tutorial,liu2020energy}. This is because the training of the classifier, when using the negative log-likelihood loss, seeks to minimize the energy of in-distribution data points. Since deriving and analyzing the energy function is not the focus of our paper, we refer the reader to the original paper~\citep{liu2020energy} for a detailed connection between the energy function and OOD detection. 

\begin{figure}[t!]
\centering
\vspace{-2cm}
\subfloat[Confidence-based Pseudo-label Decision]{%
  \includegraphics[width=0.5\linewidth]{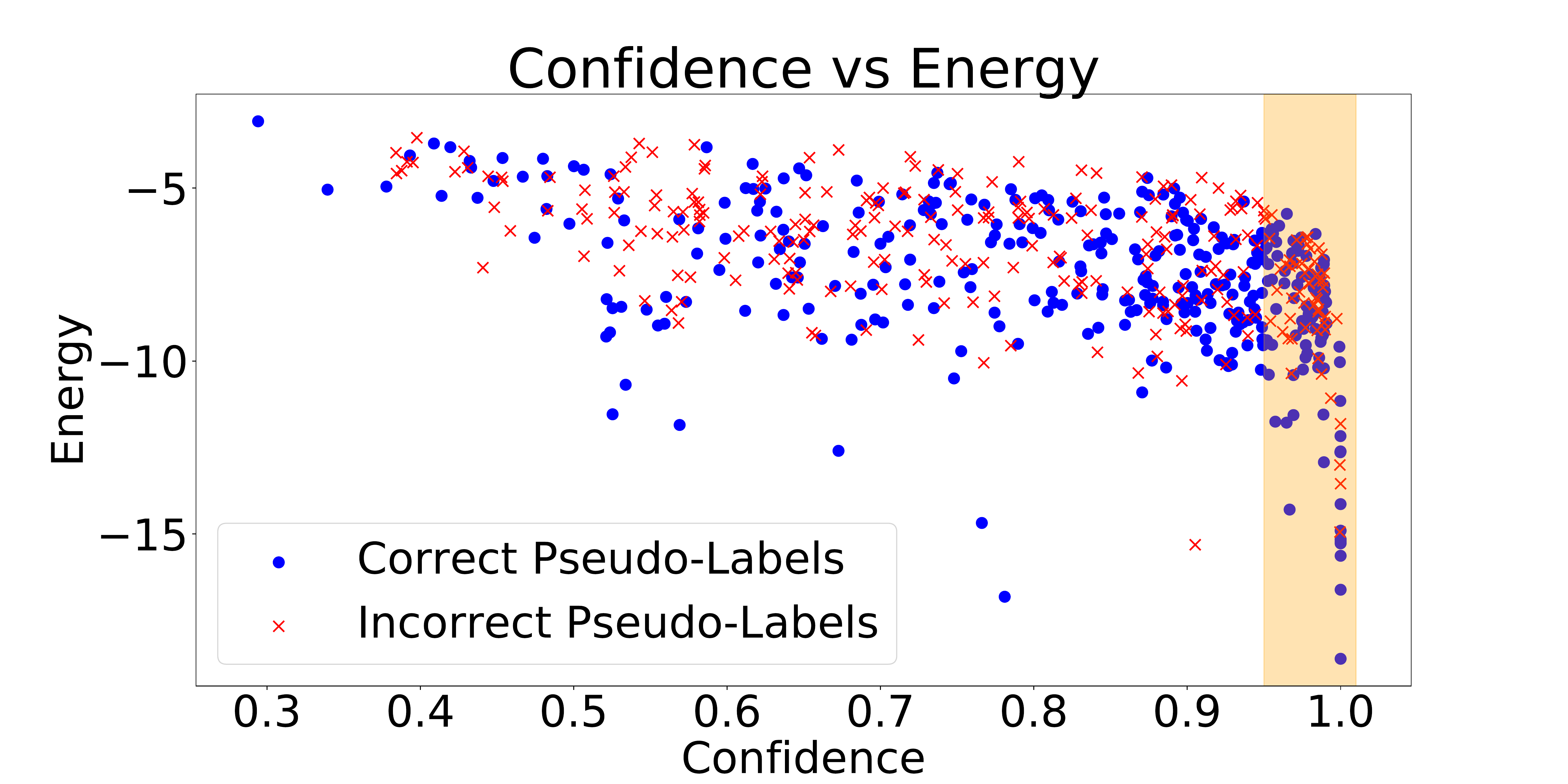}%
  \label{fig:cifar_lt_thres2}%
}
\subfloat[Energy-based Pseudo-label Decision]{%
  \includegraphics[width=0.5\linewidth]{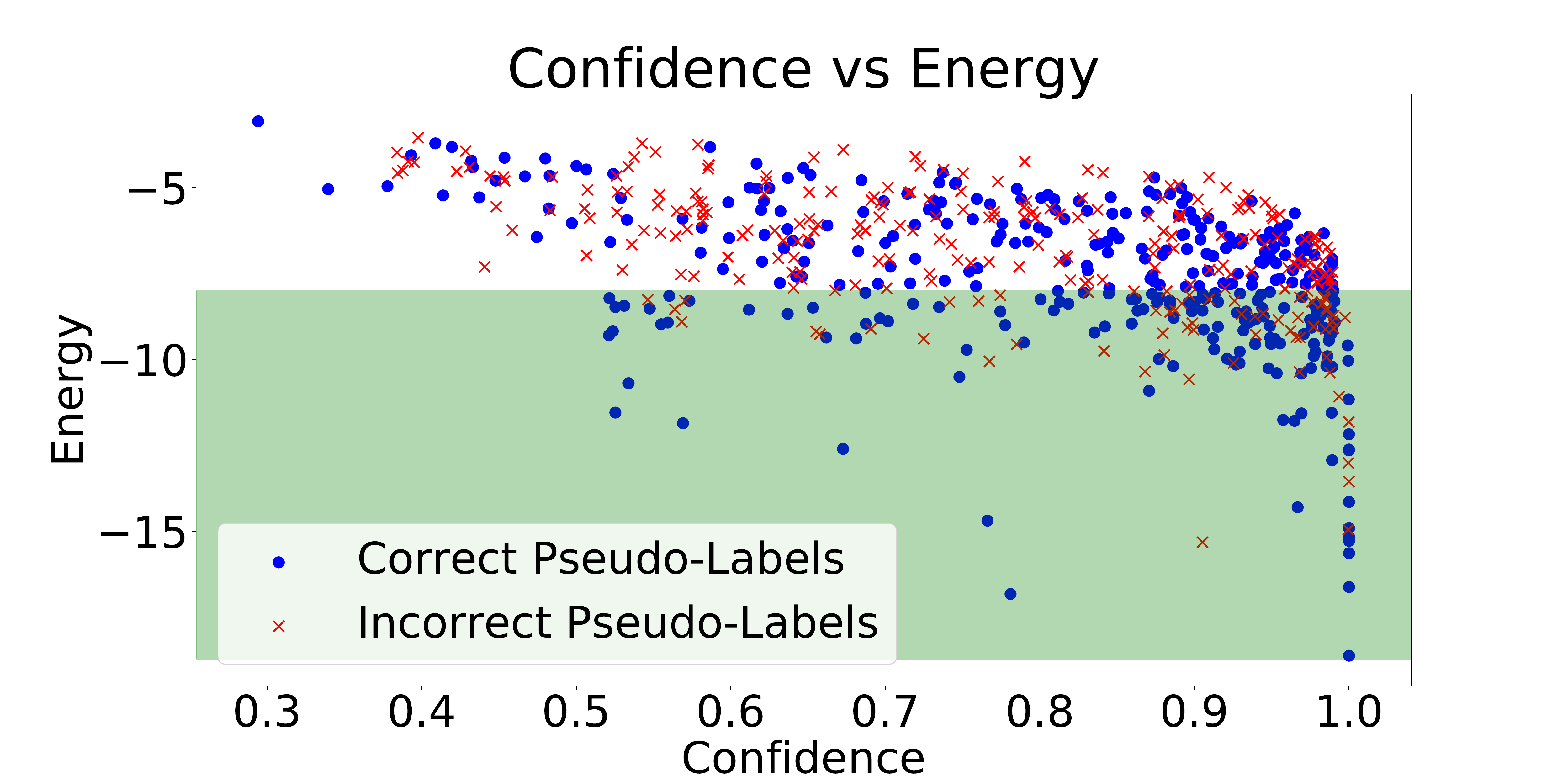}%
  \label{fig:cifar_thres2}%
}
\caption{\textbf{Visualization: confidence vs energy score}: The shaded region shows the unlabeled samples that are pseudo-labeled. Energy-based pseudo-labeling can produce correct pseudo-labels for many low-confident unlabeled samples, increasing recall while filtering out many false positives.}
\label{fig:visualization2}
\end{figure}

%

In our energy-based pseudo-labeling framework, we first compute the energy score for each unlabeled sample. We only generate a pseudo-label when the corresponding energy score is less than a pre-defined threshold $\tau_e$, which indicates that the unlabeled sample is close to the current training distribution. The actual pseudo-label is obtained by converting the model prediction on the weakly-augmented view of $\omega(\mathbf{x_b})$ to a one-hot pseudo-label. Formally, the unsupervised loss is defined as:
\begin{equation}
    \mathcal{L}_u = \frac{1}{B_u}\sum_{b=1}^{B_u} \mathbb{1} [E(\omega(\mathbf{x_b}), f(\mathbf{x_b})) < \tau_e] ~\mathcal{H}(\hat{p}(\mathbf{y}|\omega(\mathbf{x_b})), p(\mathbf{y}|\Omega(\mathbf{x_b}))).
\end{equation}

We illustrate the key difference between confidence-based pseudo-labeling and energy-based pseudo-labeling in Figure \ref{fig:overview}, and show a comparison of their sample results in Figure \ref{fig:visualization2}.  Our method can be easily integrated into SSL frameworks that use variants of confidence thresholding, without introducing additional model parameters or significant computation cost (apart from computation of the energy score). 
We shall see in the experiments that our energy-based pseudo-labeling demonstrates consistent advantages over vanilla confidence-based methods, especially in low labeled-data scenarios.
\vspace{-5pt}
\section{Experiments}
\vspace{-5pt}

We evaluate our approach on two SSL settings: (1) the imbalanced setting with long-tailed class distributions, which we believe reflects many real-world scenarios; and (2) the standard class-balanced setting, which assumes a balanced distribution of classes. Our evaluation follows standard protocols as in~\citep{oliver2018realistic, zhang2021flexmatch}.  We also conduct ablation studies to analyze important design choices of our method.

\textbf{Datasets.} We evaluate on several SSL image classification benchmarks.  For the imbalanced setting, we use CIFAR10-LT and CIFAR100-LT~\cite{krizhevsky2009learning}, which are long-tailed variants of the original CIFAR datasets.  We follow prior work in long-tail SSL~\citep{wei2021crest,lee2021abc}, and use an exponential imbalance function~\citep{cui2019class} to create the long-tailed version of CIFAR10 and CIFAR100. We select 10\% and 30\% data from each class as the labeled set for CIFAR10-LT and CIFAR100-LT, respectively. We experiment with imbalance ratio from 50 to 200 for CIFAR10-LT and from 50 to 100 for CIFAR100-LT.
Details of constructing these datasets can be found in Appendix \textcolor{red}{A.2}. 
For the standard SSL evaluation, we use balanced CIFAR10/100~\cite{krizhevsky2009learning}, whose labeled and unlabeled sets are constructed in a balanced fashion, SVHN~\cite{netzer2011reading}, and STL-10~\cite{coates2011analysis} using the standard train and test splits.

\textbf{Baselines.} We compare to the latest methods developed for long-tailed SSL (DARP~\cite{kim2020distribution}, CReST~\cite{wei2021crest}, and ABC~\cite{lee2021abc}) and for balanced SSL (UDA~\citep{xie2020unsupervised}, UPS~\citep{rizve2021defense}, FixMatch~\citep{sohn2020fixmatch}, and FlexMatch~\citep{zhang2021flexmatch}). All methods except UPS use the strong-weak data augmentation paradigm and our implementation makes the data augmentation operation consistent across different methods. FixMatch and UDA use fixed confidence thresholding with hard pseudo-labels and soft pseudo-labels, respestively. FlexMatch dynamically adjusts thresholds for different classes to (roughly) balance the pseudo-labeled samples per class. UPS adopts an uncertainty metric through MC-Dropout~\citep{gal2016dropout} by running forward pass 10 times and computing the standard deviation of outputs. This practice is very time consuming in modern SSL frameworks as the pseudo-labels are produced on-the-fly at each iteration. Therefore, for the balanced SSL setting, we report the CIFAR results from the original paper. For the imbalanced setting, we incorporate UPS into FixMatch (denoted as FixMatch-UPS) and only evaluate it on CIFAR10-LT due to its high computation cost. For SSL-LT baselines, DARP refines pseudo-labels via convex optimization designed for the imbalance scenario. CReST tries to alleviate the imbalance via moving unlabeled samples to labeled sets with probabilities inversely proportional to class frequency and restart training. ABC, in contrast, implicitly balances the classifier by introducing an auxiliary classifier trained with balanced sampling.
We implement our energy-based pseudo-labeling in the framework of FixMatch for its simplicity and denote our method as \textit{EnergyMatch}. 

\textbf{Implementation Details.} We implement our method in the open-source SSL codebase TorchSSL~\citep{zhang2021flexmatch} and conduct each experiment with three runs using different random seeds. For a fair comparison, unless otherwise specified, the baseline results and our results are generated with the same codebase, same random seeds, same data splits, and same network architecture.
Following prior work~\citep{sohn2020fixmatch, zhang2021flexmatch}, we use Wide ResNet-28-2~\citep{zagoruyko2016wide} with 1.5M parameters for Cifar10 and SVHN, WRN-28-8 for CIFAR-100, and WRN-37-2 for STL-10. All methods are trained with SGD with momentum of 0.9. For the balanced datasets, we use an initial learning rate of 0.03 and the total training iterations is set to $2^{20}$ with a cosine learning rate schedule of $\frac{7}{16}$ cycle. We use a constant learning rate for long-tailed datsets as we found using cosine decay leads to worse performance for all methods. To be consistent with our baselines~\citep{sohn2020fixmatch, zhang2021flexmatch, xie2020unsupervised}, we use a batch size of 64 for labeled data, 7x larger batch size for unlabeled data, exponential moving average with a momentum of 0.999 for inference, random horizontal flip for weak augmentation and RandAugment~\citep{cubuk2020randaugment} and CutOut~\citep{devries2017improved} for strong augmentation. Our code will be made publicly available. More training details for each experiment, hyper-parameter choices, and an anonymous link to our code base can be found in Appendix \textcolor{red}{A.1}.

\vspace{-3pt}
\subsection{Results on Long-tailed SSL}
\vspace{-3pt}

We start with experiments on the long-tailed SSL setting with imbalanced class distributions. This is a particularly challenging setting where the tail classes may only have a handful of labeled instances. As such, SSL methods that are not specifically designed for this setting can struggle to perform well.

\textbf{EnergyMatch achieves strong performance on long-tailed data}. First, we evaluate EnergyMatch built upon FixMatch~\cite{sohn2020fixmatch}, where we replace its confidence-based pseudo-labeling with our energy-based pseudo-labeling. Table~\ref{tab:main_lt} presents the results. Even though our approach does not explicitly model the long-tailed distribution, it shows a significant improvement over other standard SSL methods by a large margin over FixMatch (e.g., 4-6\% absolute percentage when imbalance ratio $\gamma>50$ for CIFAR10-LT). For CIFAR100-LT, EnergyMatch reaches 50.36\% and 44.51\% average accuracy when $\gamma=50$ and $\gamma=100$ respectively, which are also good improvements over other methods. Although FlexMatch~\citep{zhang2021flexmatch} achieves state-of-the-art results on balanced CIFAR10/100 under most settings (as we will see in Sec.\ \ref{sec:balanced}), its performance on imbalanced data is either similar to or even worse than FixMatch~\citep{sohn2020fixmatch} because it assumes both the labeled and unlabeled sets to be balanced in order to implement its flexible thresholds. Using UPS in FixMatch improves the performance over vanilla FixMatch yet still cannot match the performance of EnergyMatch on CIFAR10-LT. Moreover, UPS requires forwarding unlabeled data 10 times to compute the uncertainty measurement, which is extremely expensive in modern SSL frameworks such as FixMatch. In comparison to FixMatch+UPS, our EnergyMatch provides strong empirical results on long-tailed data and remains highly efficient. 





\begin{table}[t]
    \centering
    \caption{Top-1 accuracy on long-tailed CIFAR10/100. We use 10\% of data as labeled sets for CIFAR10-LT and 30\% data as labeled sets for CIFAR100-LT. Results are reported with the mean and standard deviation over 3 different runs.}
    \begin{adjustbox}{width=0.75\columnwidth,center}
    \begin{tabular}{lccccc}
    \toprule
        & \multicolumn{3}{c}{CIFAR10-LT} & \multicolumn{2}{c}{CIFAR100-LT} \\
        \cmidrule(l{3pt}r{3pt}){1-1} \cmidrule(l{3pt}r{3pt}){2-4}  \cmidrule(l{3pt}r{3pt}){5-6}
        Imbalance Ratio & $\gamma=50$ & $\gamma=100$ & $\gamma=200$ & $\gamma=50$ & $\gamma=100$  \\
        \cmidrule(l{3pt}r{3pt}){1-1} \cmidrule(l{3pt}r{3pt}){2-4}  \cmidrule(l{3pt}r{3pt}){5-6}
        UDA~\cite{xie2020unsupervised} & 80.81{\scriptsize $\pm$0.51} & 71.14{\scriptsize $\pm$1.98} & 62.47{\scriptsize $\pm$0.73} & 48.91{\scriptsize $\pm$0.76} & 43.11{\scriptsize $\pm$0.97}  \\
        FixMatch~\cite{sohn2020fixmatch}  & 81.54{\scriptsize $\pm$0.78} & 72.57{\scriptsize $\pm$1.37} & 62.91{\scriptsize $\pm$1.04}  & 48.97{\scriptsize $\pm$0.77} &  43.35{\scriptsize $\pm$0.95} \\
        FixMatch-UPS~\cite{rizve2021defense} & 82.45{\scriptsize $\pm$0.56} & 73.17{\scriptsize $\pm$1.63} & 65.17{\scriptsize $\pm$0.66} & - &  - \\
        FlexMatch~\cite{zhang2021flexmatch} & 79.98{\scriptsize $\pm$0.95} & 70.63{\scriptsize $\pm$1.44} &  60.87{\scriptsize $\pm$1.36} & 49.52{\scriptsize $\pm$0.48} & 43.41{\scriptsize $\pm$0.36}  \\
        \cmidrule(l{3pt}r{3pt}){1-1} \cmidrule(l{3pt}r{3pt}){2-4}  \cmidrule(l{3pt}r{3pt}){5-6}
        EnergyMatch (ours) & \textbf{83.88}{\scriptsize $\pm$0.66} & \textbf{76.81}{\scriptsize $\pm$2.08} & \textbf{67.05}{\scriptsize $\pm$1.38} & \textbf{50.36}{\scriptsize $\pm$0.88} & \textbf{44.51}{\scriptsize $\pm$0.38} \\
        \bottomrule
    \end{tabular}
    \end{adjustbox}
    
    \label{tab:main_lt}
\end{table}

\begin{table}[t!]
\centering
\caption{Top-1 accuracy on long-tailed CIFAR10/100 compared with SSL-LT methods. Following ABC~\cite{lee2021abc}, we use 20\% labeled data for CIFAR10-LT and 40\% labeled data for CIFAR100-LT. We report both the overall accuracy (before ``/'') and the accuracy of minority classes (after ``/'').}
\label{tab:abc}
\vspace{-5pt}
\begin{adjustbox}{width=\columnwidth,center}
\begin{tabular}{lcccc}
\toprule Dataset & \multicolumn{3}{c}{CIFAR10-LT}& \multicolumn{1}{c}{CIFAR100-LT} \\ 
\cmidrule(r){1-1}\cmidrule(lr){2-4}\cmidrule(lr){5-5}
\cmidrule(r){1-1}\cmidrule(lr){2-4}\cmidrule(lr){5-5}
Imbalance Ratio & $\gamma = 100$ & $\gamma = 150$ & $\gamma = 200$  & $\gamma = 20$\\
\cmidrule(r){1-1}\cmidrule(lr){2-4}\cmidrule(lr){5-5}

FixMatch~\cite{sohn2020fixmatch} & 72.3{\scriptsize $\pm$0.33} /~53.8{\scriptsize $\pm$0.63}  & 68.5{\scriptsize $\pm$0.60} /~45.8{\scriptsize $\pm$1.15} & 66.3{\scriptsize $\pm$0.49} /~42.4{\scriptsize $\pm$0.94} & 51.0{\scriptsize $\pm$0.20} /~32.8{\scriptsize $\pm$0.41}  \\
 ~w/ DARP+cRT~\cite{kim2020distribution} & 78.1{\scriptsize $\pm$0.89} /~66.6{\scriptsize $\pm$1.55} & 73.2{\scriptsize $\pm$0.85} /~57.1{\scriptsize $\pm$1.13}
           & - & 54.7{\scriptsize $\pm$0.46} /~41.2{\scriptsize $\pm$0.42}  \\
 ~w/ CReST+~\cite{wei2021crest}  & 76.6{\scriptsize $\pm$0.46} /~61.4{\scriptsize $\pm$0.85} &  70.0{\scriptsize $\pm$0.82} /~49.4{\scriptsize $\pm$1.52}
            & - & 51.6{\scriptsize $\pm$0.29} /~36.4{\scriptsize $\pm$0.46}    \\
 ~w/ ABC~\cite{lee2021abc} & 81.1{\scriptsize $\pm$0.82} /~72.0{\scriptsize $\pm$1.77}  & 77.1{\scriptsize $\pm$0.46} /~64.4{\scriptsize $\pm$0.92} & 73.9{\scriptsize $\pm$1.18} /~58.1{\scriptsize $\pm$2.72}  
            & 56.3{\scriptsize $\pm$0.19} / ~43.4{\scriptsize $\pm$0.42}   \\

 ~w/ ABC-Energy (ours) & \textbf{81.5}{\scriptsize $\pm$0.61} /~\textbf{74.5}{\scriptsize $\pm$1.47}  & \textbf{78.2}{\scriptsize $\pm$1.10} /~\textbf{66.1}{\scriptsize $\pm$2.78} 
 & \textbf{75.2}{\scriptsize $\pm$1.25} /~\textbf{60.5}{\scriptsize $\pm$2.73}& \textbf{57.0}{\scriptsize $\pm$0.41} /~\textbf{44.9}{\scriptsize $\pm$0.45}   \\ \midrule
 
 RemixMatch~\cite{berthelot2019remixmatch} & 73.7{\scriptsize $\pm$0.39} /~55.9{\scriptsize $\pm$0.87}  & 69.9{\scriptsize $\pm$0.23} /~48.4{\scriptsize $\pm$0.60} & 68.2{\scriptsize $\pm$0.37} /~45.4{\scriptsize $\pm$0.70} & 54.0{\scriptsize $\pm$0.29} /~37.1{\scriptsize $\pm$0.37}  \\
 ~w/ DARP+cRT~\cite{kim2020distribution}  & 78.5{\scriptsize $\pm$0.61} /~66.4{\scriptsize $\pm$1.69} & 73.9{\scriptsize $\pm$0.59} /~57.4{\scriptsize $\pm$1.45}
           & - & 55.1{\scriptsize $\pm$0.45} /~43.6{\scriptsize $\pm$0.58}  \\
 ~w/ CReST+~\cite{wei2021crest}   & 75.7{\scriptsize $\pm$0.34} /~59.6{\scriptsize $\pm$0.76} &  71.3{\scriptsize $\pm$0.77} /~50.8{\scriptsize $\pm$1.56}
            & - & 54.6{\scriptsize $\pm$0.48} /~38.1{\scriptsize $\pm$0.69}    \\
 ~w/ ABC~\cite{lee2021abc}  & 82.4{\scriptsize $\pm$0.45} /~75.7{\scriptsize $\pm$1.18}  & 80.6{\scriptsize $\pm$0.66} /~72.1{\scriptsize $\pm$1.51} &  78.8{\scriptsize $\pm$0.27} /~69.9{\scriptsize $\pm$0.99} 
            & \textbf{57.6}{\scriptsize $\pm$0.26} /~46.7{\scriptsize $\pm$0.50}   \\

 ~w/ ABC-Energy (ours) & \textbf{83.1}{\scriptsize $\pm$0.88} /~\textbf{77.1}{\scriptsize $\pm$1.33} & \textbf{80.8}{\scriptsize $\pm$1.03} /~\textbf{72.7}{\scriptsize $\pm$2.04} & \textbf{78.9}{\scriptsize $\pm$1.14} /~\textbf{70.3}{\scriptsize $\pm$1.55} & 57.5{\scriptsize $\pm$0.56} /~\textbf{47.8}{\scriptsize $\pm$0.64}   \\ 

\bottomrule
\end{tabular}
\end{adjustbox}
\end{table}

\textbf{Energy-based pseudo-labeling benefits methods developed for long-tailed SSL}. Next, we show that the energy-based pseudo-labeling can be readily integrated into state-of-the-art methods developed specifically for long-tailed SSL, and achieve further improvement. For this, we replace the confidence-based pseudo-labeling in the recent ABC~\citep{lee2021abc} framework with our energy-based pseudo-labeling (see imp.~details in Appendix \textcolor{red}{A.3}). We denote our method here as \textit{ABC-Energy}.

Table~\ref{tab:abc} compares our method to state-of-the-art works on long-tailed SSL. Using energy-based pseudo-labeling consistently outperforms the confidence-based counterparts on CIFAR10-LT and CIFAR100-LT. For FixMatch base, when the imbalance ratio $\gamma > 100$ on CIFAR10-LT, energy-based pseudo-labeling shows a 1.1\% and 1.3\% accuracy improvement, respectively (FixMatch w/ ABC vs. FixMatch w/ ABC-Energy). For RemixMatch base, energy-based pseudo-labeling still consistently outperforms confidence-based methods. More importantly, ABC-Energy achieves significantly higher accuracy for minority classes across all experiment protocols (numbers reported after the ``/'' in Table~\ref{tab:abc}). This indicates that our ``in-distribution vs out-of-distribution'' perspective leads to more reliable pseudo-labels when labeled data is scarce, compared to confidence-based selection. 

\vspace{-2pt}
\subsection{Results on Balanced SSL}\label{sec:balanced}
\vspace{-2pt}

We next evaluate on standard class-balanced SSL settings and use FixMatch~\citep{sohn2020fixmatch} framework again.

\textbf{EnergyMatch also produces competitive results on balanced data}. 
Table \ref{tab:main} summarizes the results. With a small ammount of labeled data (CIFAR $N=40$ and SVHN $N=40$), our EnergyMatch shows a noticeable improvement over the FixMatch baseline (+2.05\% and +1.59\% in absolute accuracy, respectively). This again demonstrates the advantage of energy-based pseudo-labeling over confidence-based pseudo-labeling in low labeled-data regimes.  In all other settings where labeled data is abundant, our method produces comparable results to FixMatch (the gap is always within $\pm$0.5\%). This is expected as most of the data points with high confidence would now be in-distribution. 
FlexMatch performs the best on the CIFAR datasets, but performs relatively poorly on SVHN because SVHN is slightly imbalanced with an imbalance ratio 2.67. This demonstrates that FlexMatch is sensitive to imbalance data even if the ratio is low. UPS, though reported with a different network backbone, consistently performs the worst among all methods.

The core idea behind existing SSL-LT methods is to combine balanced-sampling with standard SSL methods either explicitly (as CReST~\citep{wei2021crest}) or implicitly (as ABC~\citep{lee2021abc}). Thus, when the dataset is already balanced, these methods conceptually boil down to standard SSL methods such as FixMatch, and empirically also show no improvement. In contrast, our method not only achieves better results in imbalanced scenarios but also shows improvement over FixMatch in low-label regimes. 
Overall, these results, combined with the imbalanced setting results, demonstrate that our simple energy-based pseudo-labeling approach produces strong performance on both the balanced and imbalanced settings, unlike prior SSL methods which perform well in either setting, but not both. 


\begin{table}[t!]
\centering
\caption{Top-1 accuracy on CIFAR-10/100, SVHN, and STL-10. $\dagger$Due to adaptation difficulties, we report the results of UPS from its original paper~\cite{rizve2021defense}, which uses a different network backbone. All other methods (including ours) share the same backbone across experiments. For each experiment, we bold the best result and underline the second best result.}
\label{tab:main}
\begin{adjustbox}{width=\columnwidth,center}
\begin{tabular}{@{}lcccccccccc@{}}
\toprule  & \multicolumn{3}{c}{CIFAR10}& \multicolumn{3}{c}{CIFAR100}& \multicolumn{2}{c}{SVHN} & \multicolumn{1}{c}{STL-10} \\ \cmidrule(lr){2-4}\cmidrule(lr){5-7}\cmidrule(lr){8-9}\cmidrule(l){10-10}
 \, & \multicolumn{1}{c}{$N=40$} & \multicolumn{1}{c}{$N=250$}  & \multicolumn{1}{c}{$N=4000$} & \multicolumn{1}{c}{$N=400$}  & \multicolumn{1}{c}{$N=2500$}  & \multicolumn{1}{c}{$N=10000$} & \multicolumn{1}{c}{$N=40$}  & \multicolumn{1}{c}{$N=250$}  &  \multicolumn{1}{c}{$N=1000$} \\ 
 \cmidrule(r){1-1}\cmidrule(lr){2-4}\cmidrule(lr){5-7}\cmidrule(lr){8-9} \cmidrule(l){10-10}
 \,UPS$\dagger$~\cite{rizve2021defense} & -  & - & 93.61{\scriptsize $\pm$0.02} &  -  & -  & 68.00{\scriptsize $\pm$0.49} & -  & - & -\\
 
 \,UDA~\cite{xie2020unsupervised}    & 89.38{\scriptsize $\pm$3.75}  & 94.84{\scriptsize $\pm$0.06} & 95.71{\scriptsize $\pm$0.07} &  \underline{53.61}{\scriptsize $\pm$1.59}  & \underline{72.27}{\scriptsize $\pm$0.21}  & 77.51{\scriptsize $\pm$0.23} & 94.88{\scriptsize $\pm$4.27}  &  \textbf{98.08}{\scriptsize $\pm$0.05} &  93.36{\scriptsize $\pm$0.17}\\

\,FixMatch~\cite{sohn2020fixmatch} & 92.53{\scriptsize $\pm$0.28}  & \textbf{95.14}{\scriptsize $\pm$0.05} & \underline{95.79}{\scriptsize $\pm$0.08}  & 53.58{\scriptsize $\pm$0.82} & 71.97{\scriptsize $\pm$0.16} & \underline{77.80}{\scriptsize $\pm$0.12} & \underline{96.19}{\scriptsize $\pm$1.18}  & \underline{97.98}{\scriptsize $\pm$0.02} & 93.75{\scriptsize $\pm$0.03} \\

\,FlexMatch~\cite{zhang2021flexmatch} & \textbf{95.03}{\scriptsize $\pm$0.06} & \underline{95.02}{\scriptsize $\pm$0.09} & \textbf{95.81}{\scriptsize $\pm$0.01} & \textbf{60.06}{\scriptsize $\pm$1.62} & \textbf{73.51}{\scriptsize $\pm$0.20} & \textbf{78.10}{\scriptsize $\pm$0.15} & 91.81{\scriptsize $\pm$3.02} & 93.41{\scriptsize $\pm$2.29} & \textbf{94.23}{\scriptsize $\pm$0.18} \\ 
 \cmidrule(r){1-1}\cmidrule(lr){2-4}\cmidrule(lr){5-7}\cmidrule(lr){8-9} \cmidrule(l){10-10}

\,EnergyMatch (ours) & \underline{94.58}{\scriptsize $\pm$0.43}  & 94.89{\scriptsize $\pm$0.13} & 95.72{\scriptsize $\pm$0.08} & 53.32{\scriptsize $\pm$0.57} & 71.96{\scriptsize $\pm$0.44} & 77.40{\scriptsize $\pm$0.34}  & \textbf{97.78}{\scriptsize $\pm$0.05} & 97.84{\scriptsize $\pm$0.01}  &  \underline{93.82}{\scriptsize $\pm$0.11} \\
  
\bottomrule
\end{tabular}
\end{adjustbox}
\end{table}


\begin{figure}[t]
\vspace{-15pt}
\centering
\subfloat[Threshold: CIFAR10]{%
  \includegraphics[width=.29\linewidth]{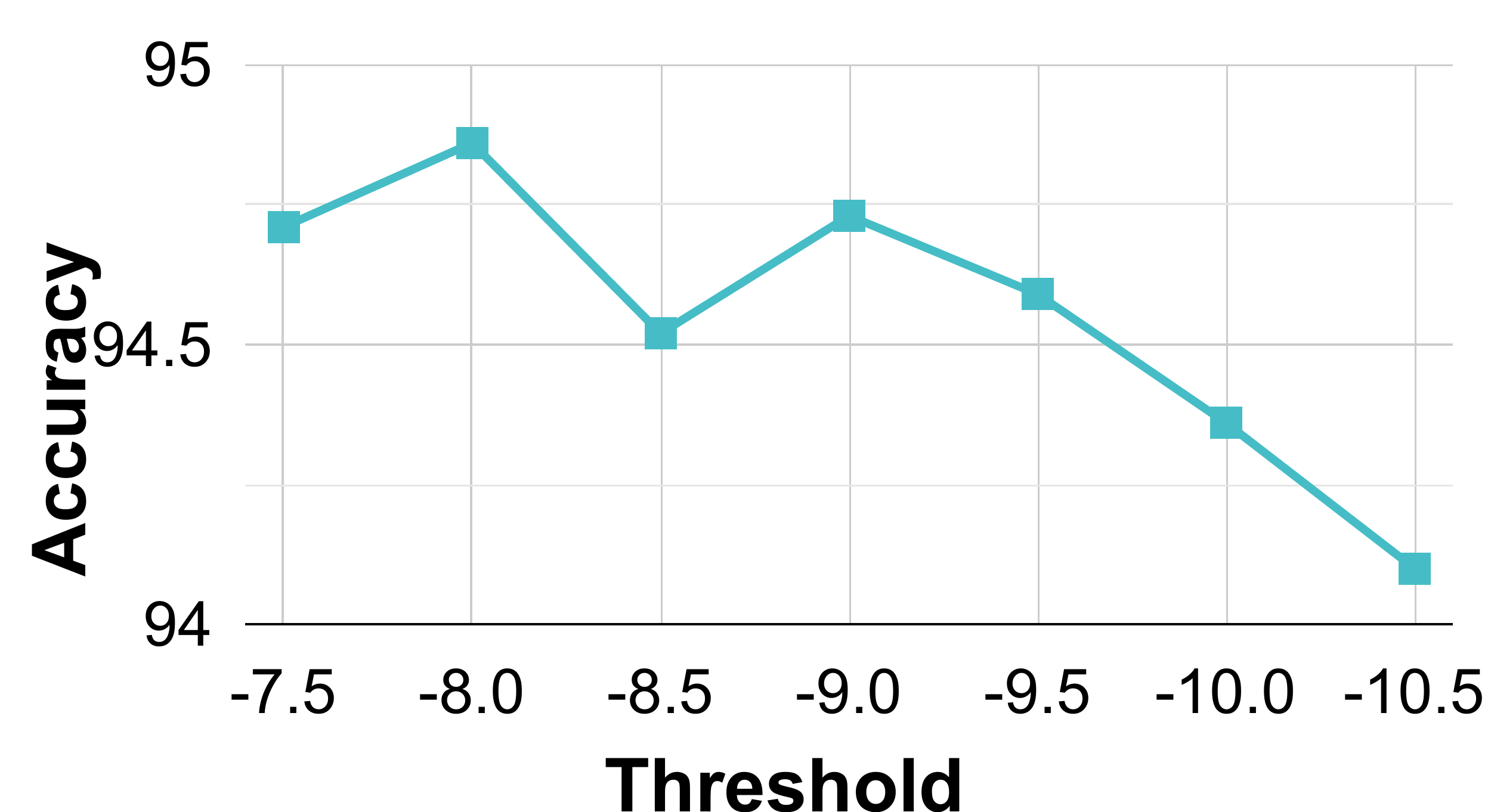}%
  \label{fig:cifar_thres}%
}\qquad
\subfloat[Threshold: CIFAR10-LT]{%
  \includegraphics[width=.29\linewidth]{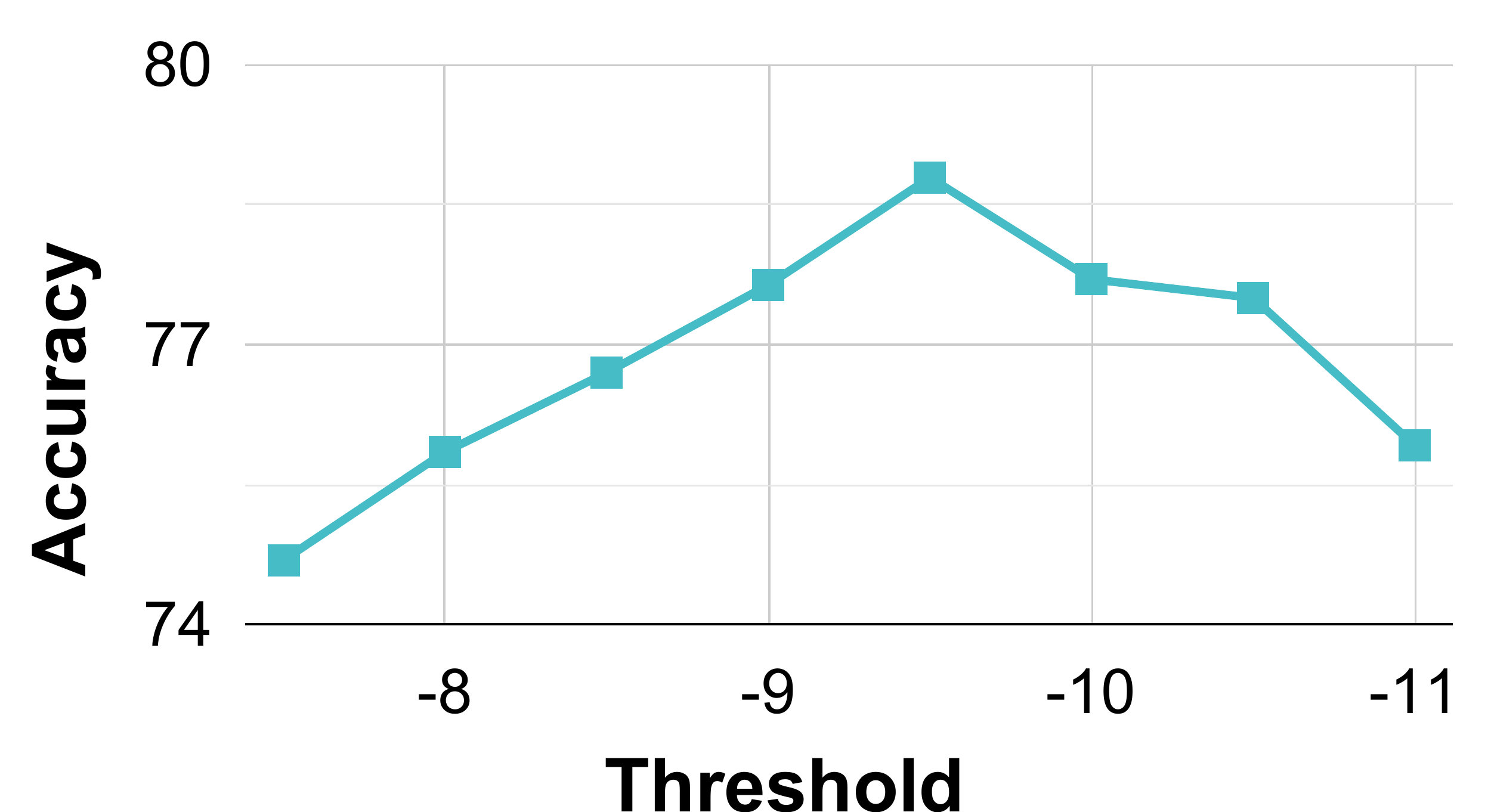}%
  \label{fig:cifar_lt_thres}%
}\qquad
\subfloat[Temperature: CIFAR10]{%
  \includegraphics[width=.29\linewidth]{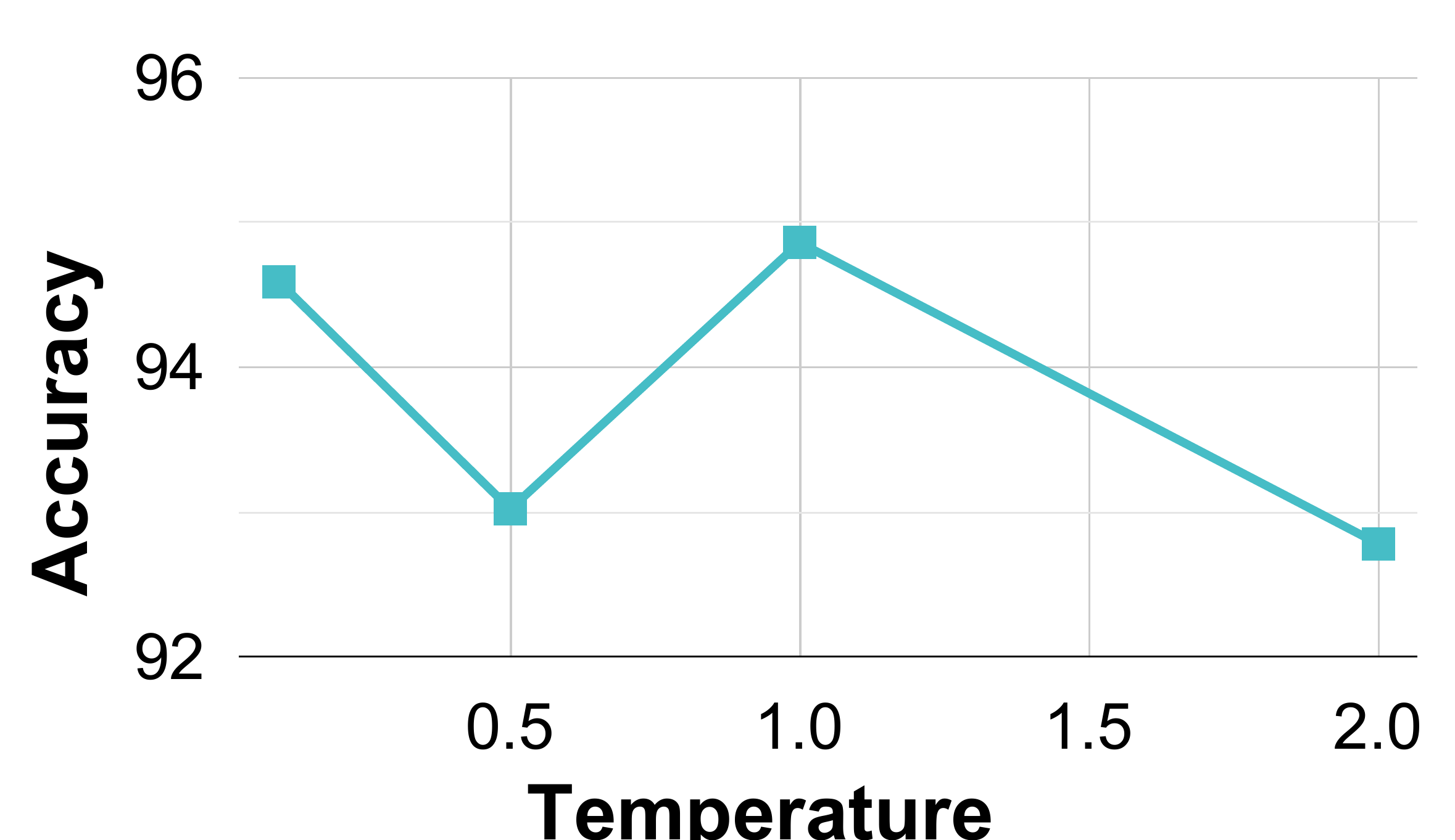}%
  \label{fig:temperature}%
}\qquad
\vspace{-5pt}
\caption{Ablation study: (a) and (b): Effects of different energy thresholds on CIFAR10 and CIFAR10-LT. (c): Effects of the temperature parameter in the energy function on CIFAR10.}
\label{fig:threshold}
\end{figure}

\vspace{-2pt}
\subsection{Ablation Studies}
\label{sec:ablation}
\vspace{-2pt}

We next conduct ablation studies to better understand our energy-based pseudo-labeling method, again integrated into the framework of FixMatch~\cite{sohn2020fixmatch}. Unless otherwise noted, experiments are conducted on CIFAR10 with 40 labels and CIFAR10-LT with imbalance ratio 100 and 10\% labels.


\textbf{Effect of energy threshold.} The most important hyper-parameter of our method is the energy threshold. Unlike confidence scores that range from 0 to 1, energy scores are unbounded with its scale proportional to the number of classes. Our thresholds in the experiments are chosen via cross-validation with a separate sampling seed (see Appendix \textcolor{red}{A.1}). We further experiment with thresholds. 
As shown in Figure~\ref{fig:threshold} (a) for balanced CIFAR10, the performance is generally similar across a threshold range of -7.5 to -9.5, and starts to decrease with lower thresholds. 
This is because with very low thresholds, the model becomes very conservative and only produces pseudo-labels for unlabeled samples that are very close to the training distribution; i.e., the recall in pseudo-labels takes a significant hit. As for long-tailed CIFAR10 (Figure~\ref{fig:threshold} (b)), a slightly lower threshold of -9.5 achieves the best performance. Since head classes have the most labeled samples, and the model could become biased to them, a stricter threshold helps to reduce the number of pseudo-labels for head classes (at a greater extent than for tail classes). See our analysis at the end of this sub-section.

\textbf{Effect of temperature for the energy score.} Recall that the energy function~\cite{lecun2006tutorial} has a tunable temperature $T$ (Equation \ref{eq:energy}). Here, we empirically evaluate the impact of this parameter. As shown in Figure \ref{fig:threshold} (c), the simplest setting of $T=1$ gives the best performance. We also experiment with a much larger temperature $T=10$, which leads to big drop in performance to 80.6\% accuracy (not shown in the plot). As noted by prior work~\cite{liu2020energy}, larger $T$ results in a smoother distribution of energy scores, which makes it harder to distinguish samples. We find that $T < 1$ also results in slightly worse performance. Therefore, we simply set $T=1$ and omit the temperature parameter to reduce the effort of hyper-parameter tuning it for our method.

\textbf{Why does our method work well on imbalanced data?}
A key advantage of energy-based pseudo-labeling is its strong performance on long-tailed data (recall Table~\ref{tab:main_lt}).  To help explain this, we provide a detailed pseudo-label precision and recall analysis. Here, we refer to the three most frequent classes as head classes, the three least frequent classes as tail classes, and the rest as body classes.

\begin{figure}[t]
\centering
\subfloat[Precision: Overall]{%
  \includegraphics[width=.25\linewidth]{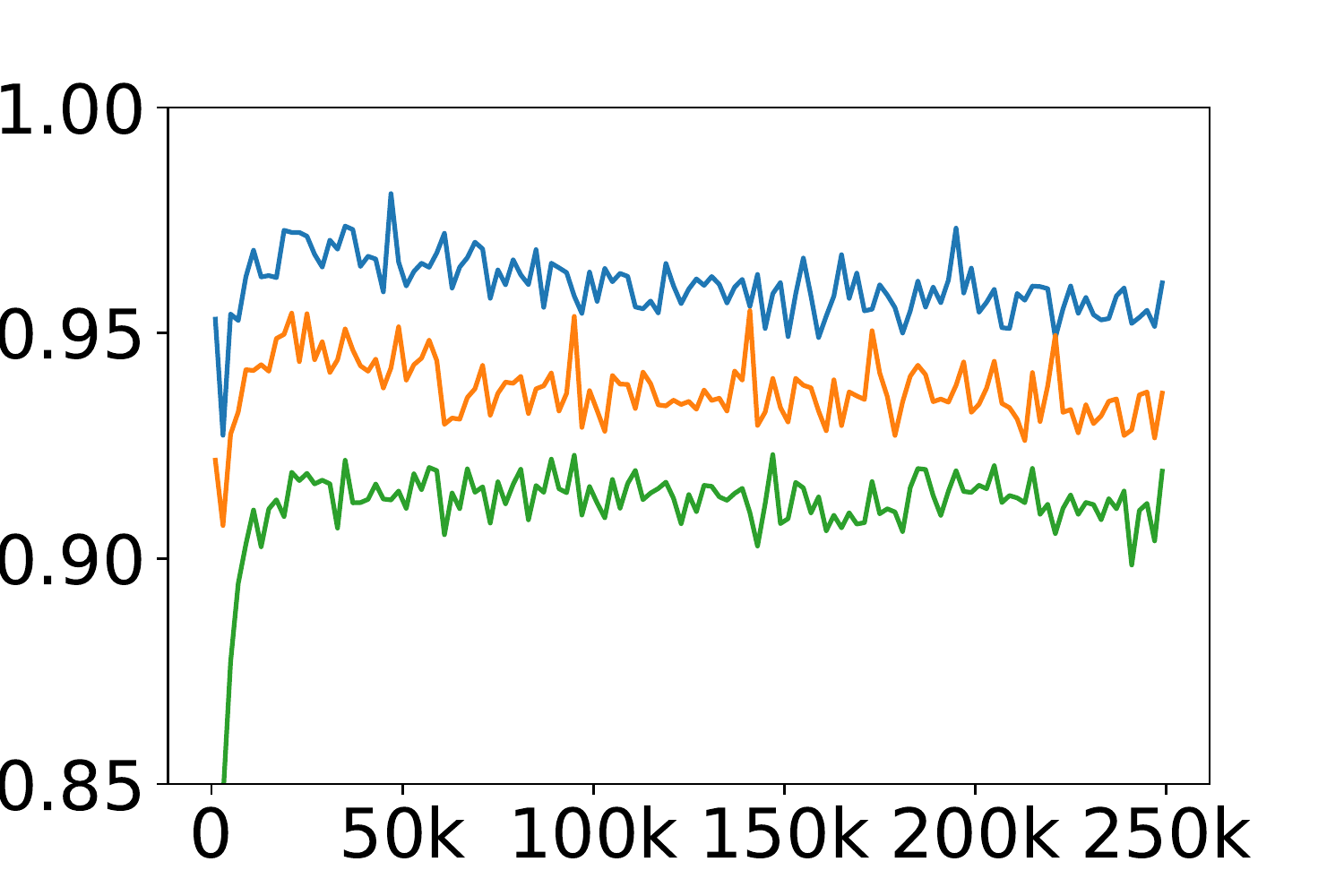}
  \label{fig:pre_overall}
}
\subfloat[Precision: Tail]{%
  \includegraphics[width=.25\linewidth]{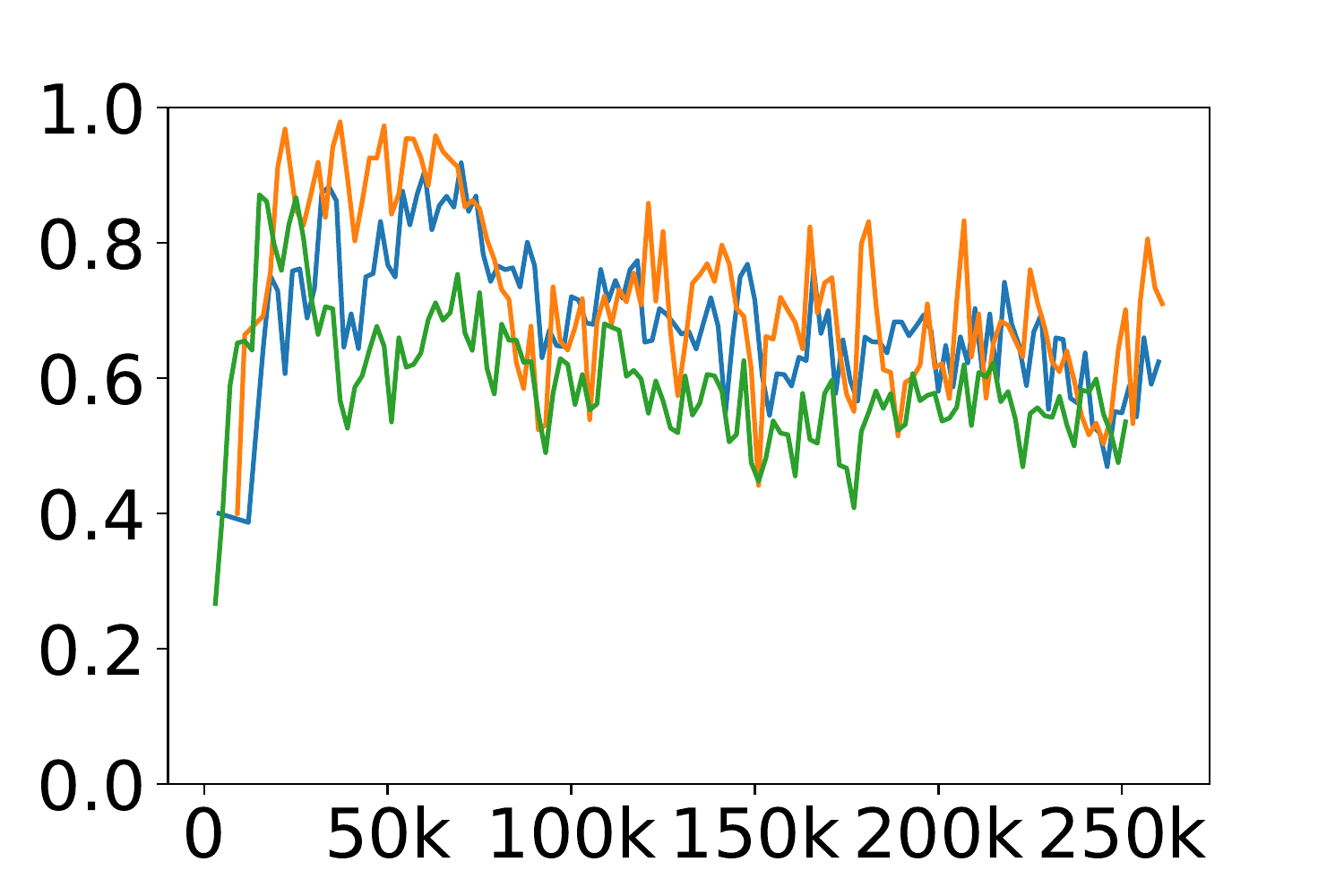}%
  \label{fig:pre_head}
}
\subfloat[Recall: Overall]{%
  \includegraphics[width=.25\linewidth]{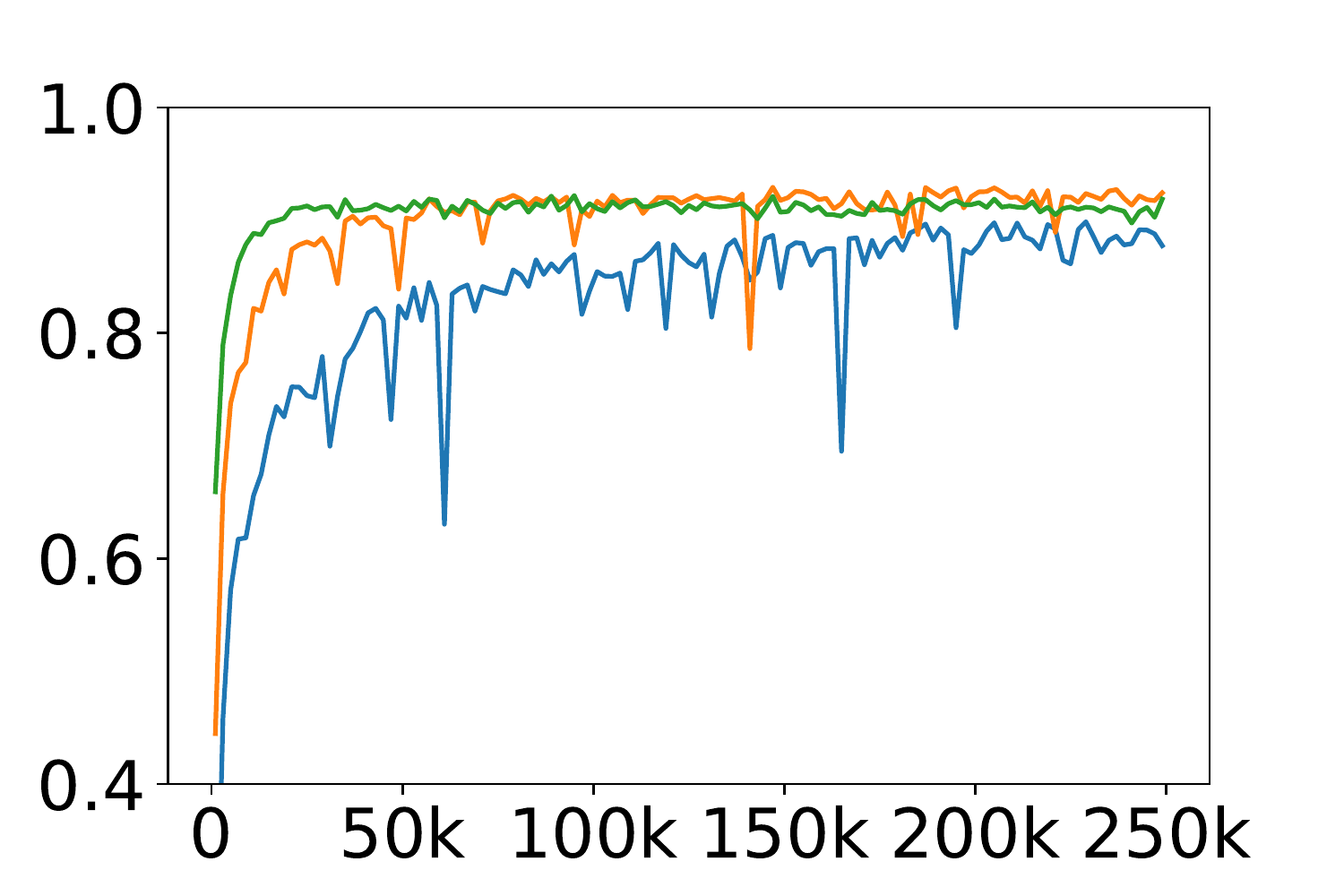}%
  \label{fig:rec_overall}
}
\subfloat[Recall: Tail]{%
  \includegraphics[width=.25\linewidth]{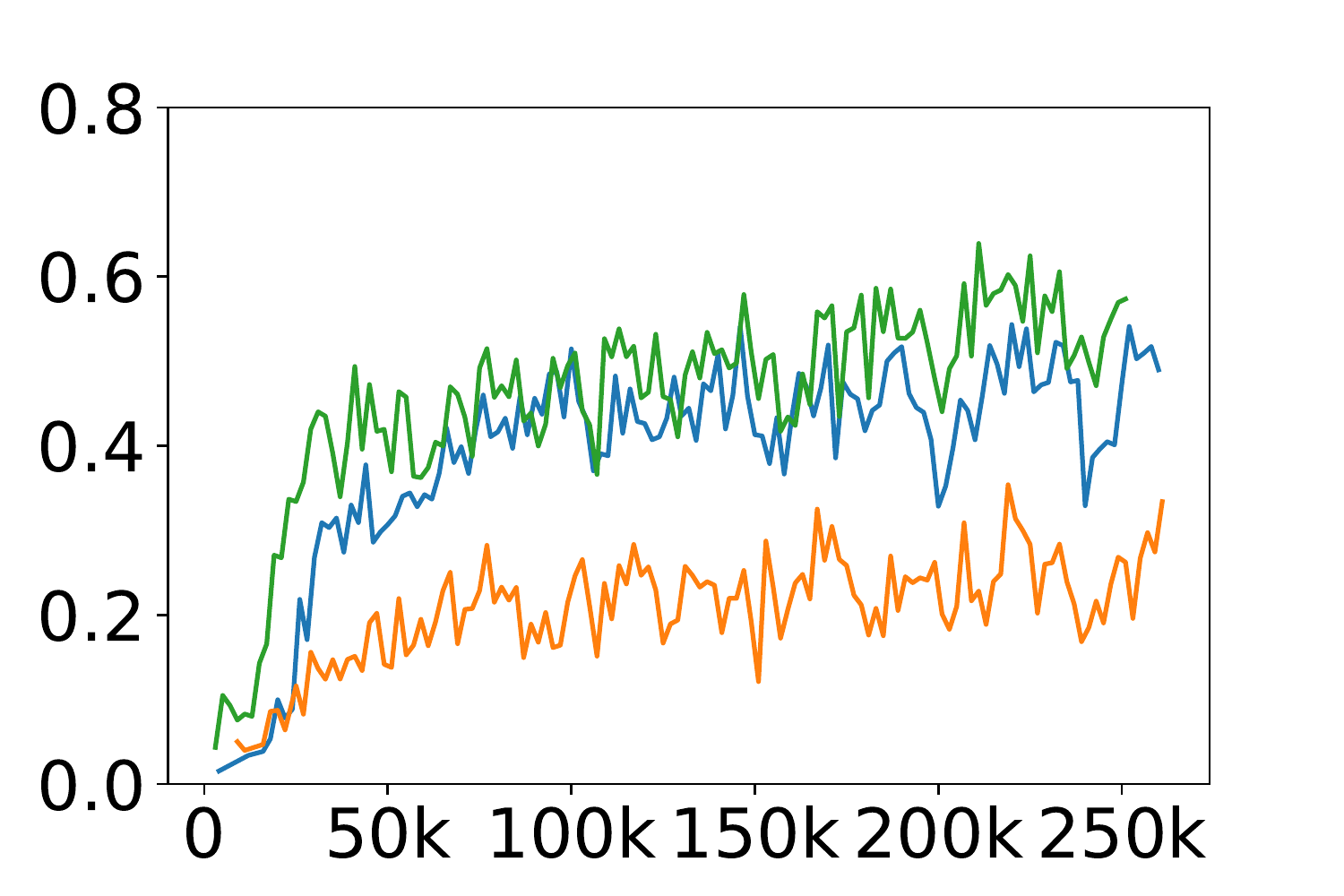}%
  \label{fig:rec_tail}
}
\vspace{-5pt}
\caption{\textbf{Precision-Recall Analysis}: We compare pseudo-label precision and recall between EnergyMatch and FixMatch. \textcolor{orange}{Orange} and \textcolor{green}{green} curves denote FixMatch with threshold 0.95 and 0.6 respectively. EnergyMatch is denoted by \textcolor{blue}{blue} curves. Although FixMatch with a lower confidence threshold (0.6) improves its recall for tail classes, its overall precision significantly drops. Our EnergyMatch achieves improved recall for tail classes and better overall precision.}
\label{fig:pr_lt}
\end{figure}
\begin{table}[t]
    \centering
    \footnotesize
    \caption{Comparison to FixMatch with various confidence thresholds on CIFAR10-LT. Results are generated with one run of 10\% labeled data and imbalance ratio 100 with the same random seed.}
    \begin{adjustbox}{width=0.87\columnwidth,center}
    \begin{tabular}{lcccccc}
    \toprule
        & \multicolumn{5}{c}{FixMatch~\cite{sohn2020fixmatch}} & \multicolumn{1}{c}{EnergyMatch (ours)} \\ 
        \cmidrule(l{3pt}r{3pt}){2-6} \cmidrule(l{3pt}r{3pt}){7-7} 
        Confidence threshold & $\tau=0.95$ & $\tau=0.9$ & $\tau=0.8$ & $\tau=0.7$ & $\tau=0.6$ & -  \\ 
        \cmidrule(l{3pt}r{3pt}){2-6} \cmidrule(l{3pt}r{3pt}){7-7} 
        Accuracy & 73.73 & 71.9 & 74.12 & 71.53 & 73.55 & \textbf{78.79}  \\
        \bottomrule
    \end{tabular}
    \end{adjustbox}
    \vspace{-10pt}
    \label{tab:fixmatch}
\end{table}

Figure \ref{fig:pr_lt} shows the precision and recall of our model's predicted pseudo-labels over all classes (a,c) and also for the tail classes (b,d). The analysis for the head and body classes can be found in Appendix \textcolor{red}{A.4}. Compared with FixMatch, our EnergyMatch achieves higher precision for overall, head, and body pseudo-labels. More importantly, it doubles FixMatch's recall of tail pseudo-labels without hurting the precision much. This shows that our model predicts more true positives for the tail classes and also becomes less biased to the head classes.  Trivially lowering the confidence threshold for FixMatch can also improve its recall for tail pseudo-labels. However, doing so significantly hurts its precision and does not improve overall accuracy. For example, as shown in Figure \ref{fig:pr_lt} (a) and (d), although using a lower threshold of 0.6 improves FixMatch's recall of tail pseudo-labels, the overall precision is significantly hurt, which results in minimal improvement in terms of overall accuracy. In Table \ref{tab:fixmatch}, we further investigate various lower confidence thresholds for FixMatch, and find that none of them leads to improvement in accuracy to match the level of EnergyMatch. 

\textbf{Other ablation studies}. We provide more analysis on pseudo-label precision and recall, and results with true out-of-distribution unlabeled data. These can be found in the Appendix.

\vspace{-5pt}
\section{Discussion and Conclusion}
\vspace{-5pt}

In this work, we presented a novel ``in-distribution vs.~out-distribution'' perspective for pseudo-labeling in SSL. Rather than making pseudo-labeling decisions based on the model's confidence, our approach instead makes that decision based on an unlabeled sample's energy score derived from the model's output. We showed that our method can be easily integrated into state-of-the-art SSL frameworks that combine pseudo-labeling with consistency regularization, and that it achieves strong results particularly in low-labeled data settings. One limitation of our method is the lack of interpretability of the energy scores; unlike confidence scores, which can be interpreted as probabilities, the energy score has a different scale and is harder to interpret. Devising ways to better understand it would be interesting future work. Overall, we believe our work has shown the promise of energy-based approaches for SSL, and hope that it will spur further research in this direction. 

In terms of societal impact, research in SSL has the potential to positively impact real-world applications that require lots of labeled data by reducing annotation effort and cost. However, it is possible that due to the automatic labeling of unlabeled data, unforeseen negative biases from the unlabeled data may creep into the model unchecked. Thus, such potential issues must be well thought out when deploying SSL models in real-world applications.




\bibliographystyle{abbrvnat}
\bibliography{reference}

\clearpage

\appendix

\setcounter{figure}{0}
\setcounter{table}{0}
\setcounter{section}{0}
\renewcommand{\theequation}{\Alph{equation}}
\renewcommand{\thefigure}{\Alph{figure}}
\renewcommand{\thesection}{\Alph{section}}
\renewcommand{\thetable}{\Alph{table}}

\section{Appendix}

This document complements the main paper by describing: (1) training details of each experiment in the main paper (Appendix \ref{app:detail}, \ref{app:cifar}, and \ref{app:abc}); (2) more precision and recall analysis for pseudo-labels (Appendix \ref{app:pr}); (3) additional experiments studying the impact of true out-of-distribution data as unlabeled data points (Appendix \ref{app:ood}); and (4) additional baseline results (Appendix \ref{app:baseline}). 

For sections, figures, tables, and equations, we use numbers (e.g., Table 1) to refer to the main paper and capital letters (e.g., Table A) to refer to this supplement.

\subsection{Training details and hyper-parameters}
\label{app:detail}

We first report the training details and hyper-parameters for reproduction of our results in Table 1 and Table 3 of the main paper; see Table \ref{tab:hyper}.
All experiments use the same set of learning rate, loss weight, batch size, and EMA momentum.  Additionally, we use SGD as the optimizer for all these results and train the model with $2^{20}$ iterations on standard SSL benchmarks and $6 \times 2^{16}$ iterations on long-tailed CIFAR10/100. 

\begin{table}[h]
    \centering
    \footnotesize
    \caption{Hyper-parameter and training details for results in Table 1 and Table 3. All experiments use the same set of learning rate, loss weight, batch size, and EMA momentum.}
    \begin{adjustbox}{width=0.87\columnwidth,center}
    \begin{tabular}{ccccccc}
    \toprule
  Hyper-parameter & CIFAR10 & SVHN  & STL10 & CIFAR100 & CIFAR10-LT & CIFAR100-LT \\ 
  \cmidrule(l{3pt}r{3pt}){1-1} \cmidrule(l{3pt}r{3pt}){2-2} \cmidrule(l{3pt}r{3pt}){3-3} \cmidrule(l{3pt}r{3pt}){4-4} \cmidrule(l{3pt}r{3pt}){5-5} \cmidrule(l{3pt}r{3pt}){6-6} \cmidrule(l{3pt}r{3pt}){7-7}
  Energy Threshold &  -8.0 & -8.0 & -8.0 & -11.0 & -9.5 & -12.5 \\
  \cmidrule(l{3pt}r{3pt}){1-1} \cmidrule(l{3pt}r{3pt}){2-2} \cmidrule(l{3pt}r{3pt}){3-3} \cmidrule(l{3pt}r{3pt}){4-4} \cmidrule(l{3pt}r{3pt}){5-5} \cmidrule(l{3pt}r{3pt}){6-6} \cmidrule(l{3pt}r{3pt}){7-7}
  Weight Decay & 0.0005 & 0.0005 & 0.0005 & 0.001 & 0.0005 & 0.001 \\
  \cmidrule(l{3pt}r{3pt}){1-1} \cmidrule(l{3pt}r{3pt}){2-2} \cmidrule(l{3pt}r{3pt}){3-3} \cmidrule(l{3pt}r{3pt}){4-4} \cmidrule(l{3pt}r{3pt}){5-5} \cmidrule(l{3pt}r{3pt}){6-6} \cmidrule(l{3pt}r{3pt}){7-7}
  Learning Rate & 0.03 & 0.03 & 0.03 & 0.03 & 0.03 & 0.03 \\
  Weight of Unsupervised Loss & 1.0 & 1.0 & 1.0 & 1.0 & 1.0 & 1.0 \\
  Labeled Batch Size & 64 & 64 & 64 & 64 & 64 & 64 \\
  Unlabeled Batch Size & 448 & 448 & 448 & 448 & 448 & 448 \\
  EMA momentum & 0.999 & 0.999 & 0.999 & 0.999 & 0.999 & 0.999 \\

  \bottomrule
  
  \end{tabular}
    \end{adjustbox}
    
    \label{tab:hyper}
\end{table}

\subsection{Construction of Long-tailed CIFAR}
\label{app:cifar}
In this section, we describe the construction of the long-tailed version of CIFAR10/100 for semi-supervised learning. First, we construct a long-tailed version of these datasets using all the labels. Specifically, with imbalance ratio $\gamma$, the number of labels for class $k$ is computed as $N_k = N_1 \cdot \gamma ^ {-(k-1)/(K-1)}$, where $K$ is the total number of classes and $N_1$ is the number of labels for the most frequent class. 
For CIFAR10, $N_1 = 5000$ and $K = 10$. For CIFAR100, $N_1 = 500$ and $K = 100$. After constructing the long-tailed version of each dataset, we randomly sample a certain percentage of data with labels from each class evenly as the labeled set and use the rest as unlabeled set so that both labeled and unlabeled set follow the same long-tailed distribution.  For results in Table 1, we use 10\% of data from each class as the labeled set of CIFAR10 and use 30\% of data from each class as labeled set of CIFAR100. 

\subsection{Training Details of ABC}
\label{app:abc}

For results in Table 2, we follow the training settings of ABC~\cite{lee2021abc}. Specifically, for all the results, we use ADAM~\cite{kingma2014adam} as the optimizer with a constant learning rate 0.002 and train models with 25000 iterations. 
Exponential moving average of momentum 0.999 is used to maintain an ensemble network for evaluation. The strong augmentation transforms include RandAugment and CutOut, which is consistent with the standard practice in SSL. Batch size of both labeled set and unlabeled set is set to 64 and two strongly-augmented views of each unlabeled sample are included in training. Following ABC, WideResNet-28-2 is used for both CIFAR10 and CIFAR100 for results in Table 2.

\subsection{Precision and Recall for Head and Body Pseudo-labels}
\label{app:pr}

\begin{figure}[t]
\centering
\subfloat[Precision: Head]{%
  \includegraphics[width=.25\linewidth]{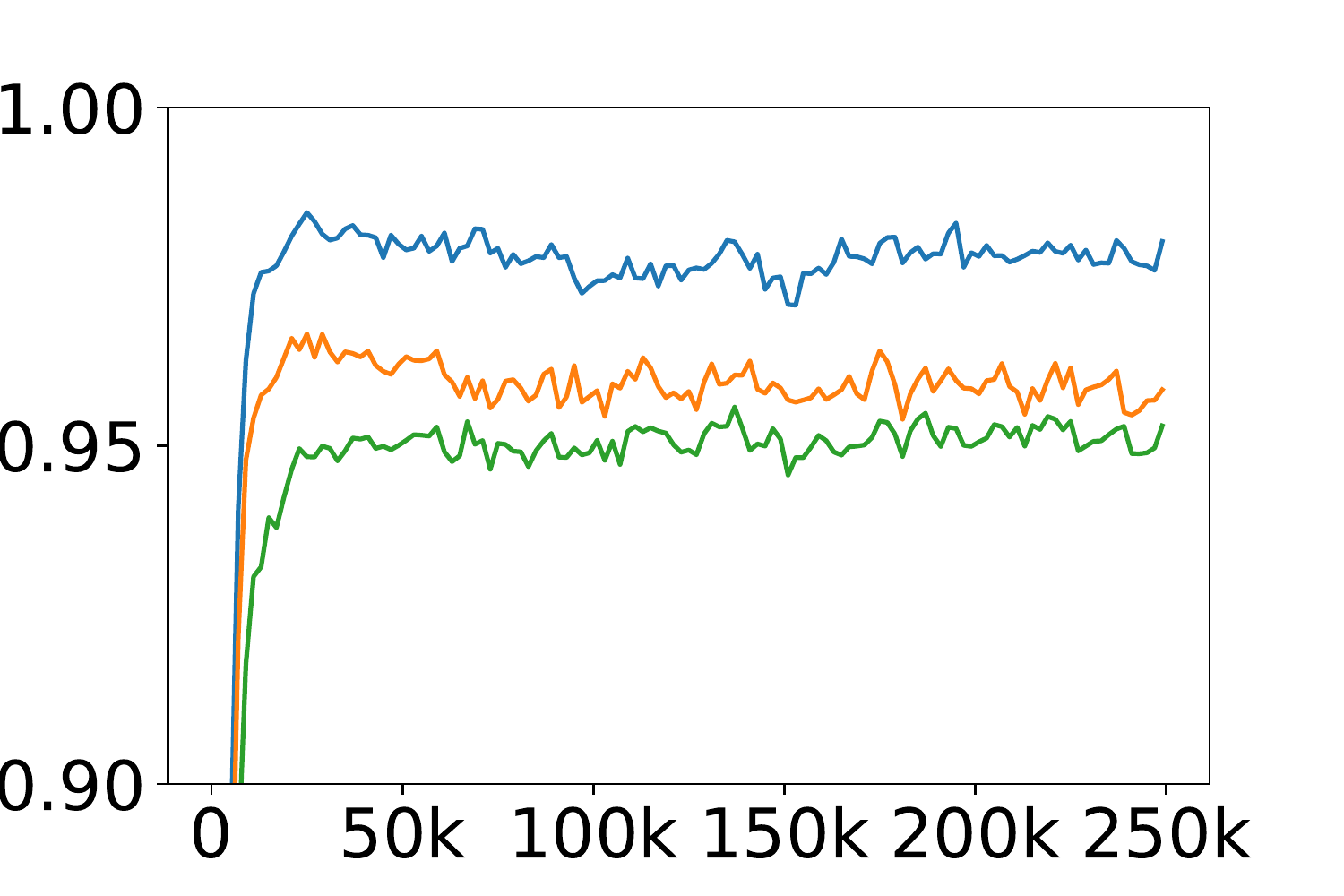}
  \label{fig:pre_headl}
}
\subfloat[Precision: Body]{%
  \includegraphics[width=.25\linewidth]{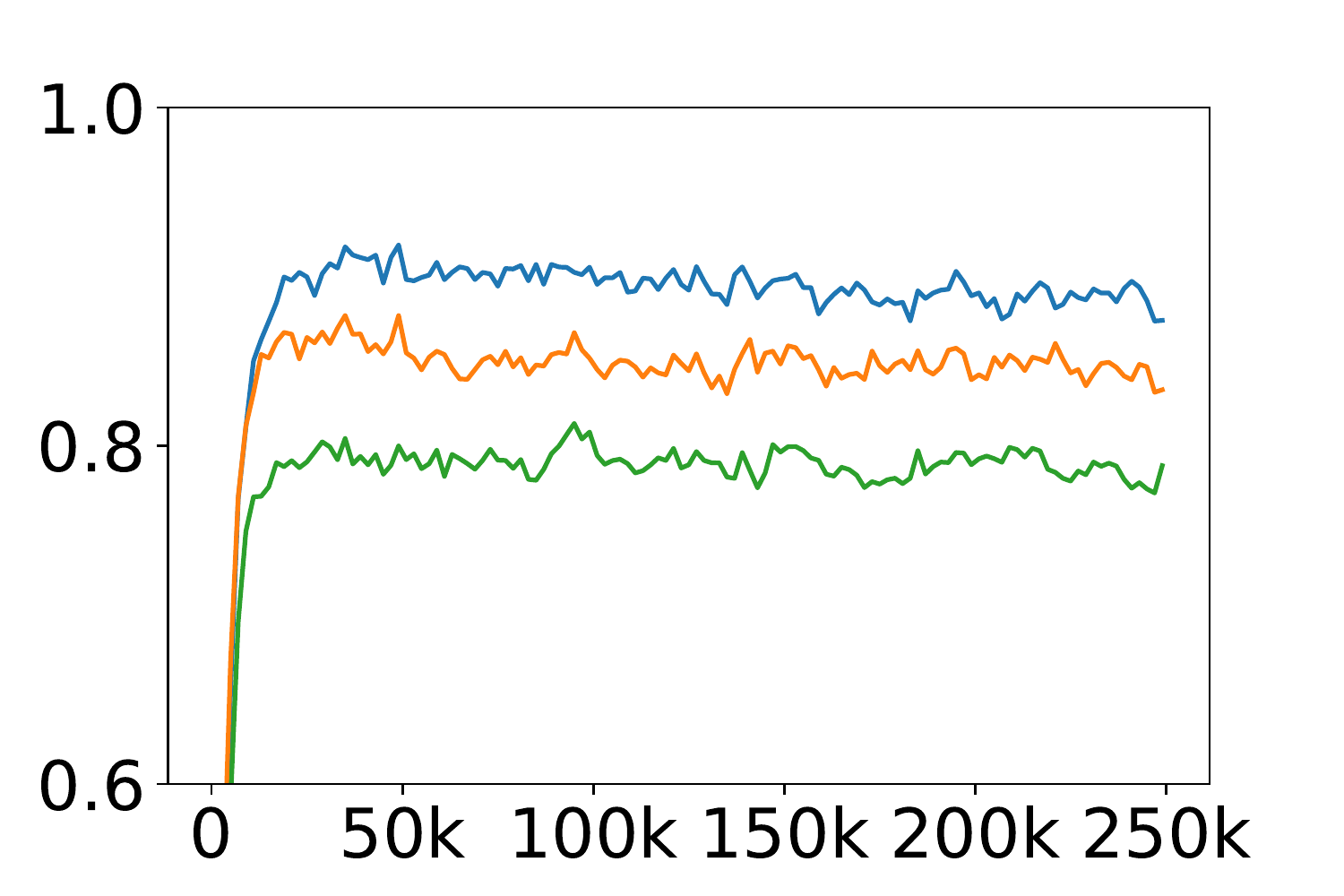}%
  \label{fig:pre_body}
}
\subfloat[Recall: Head]{%
  \includegraphics[width=.25\linewidth]{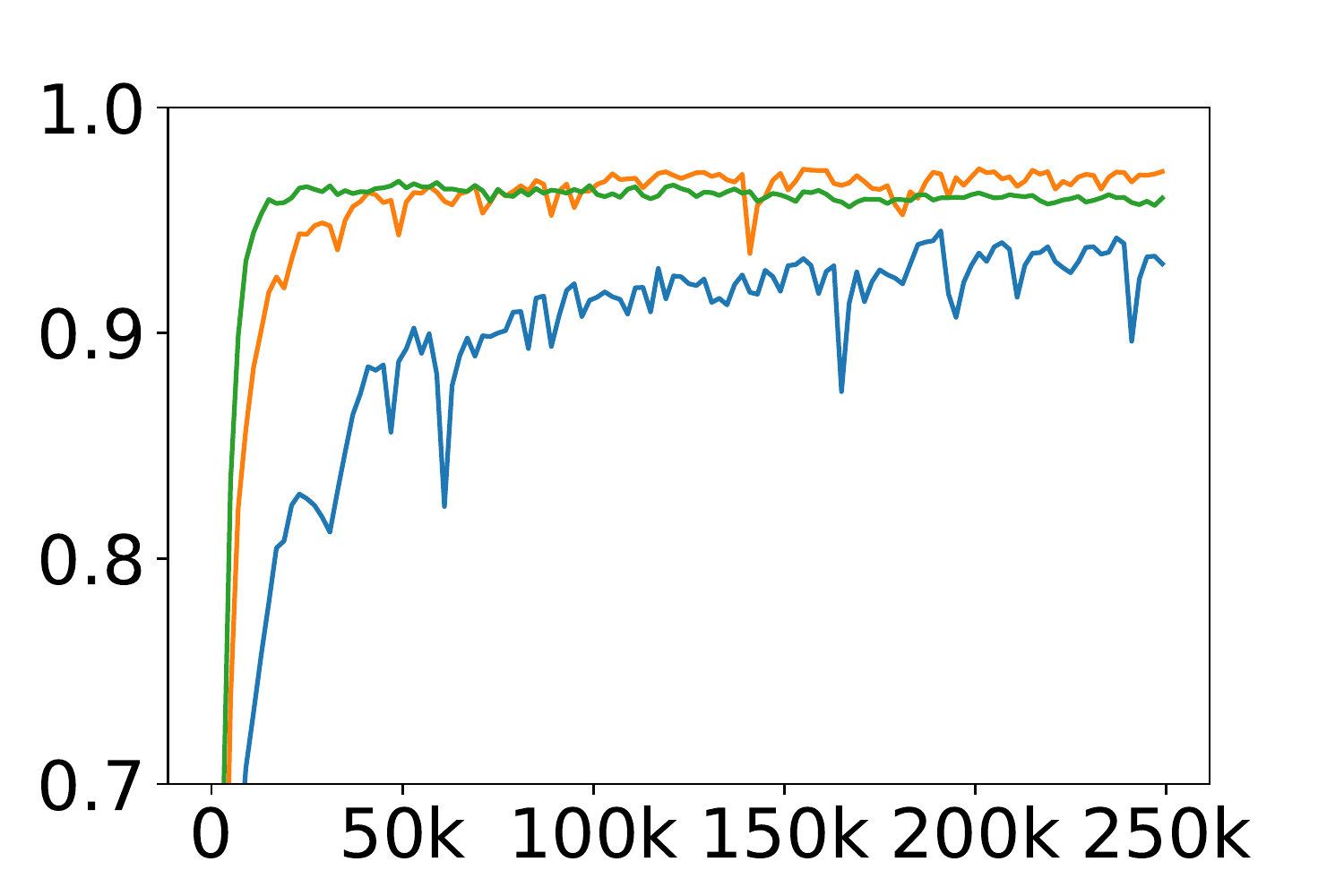}%
  \label{fig:rec_head}
}
\subfloat[Recall: Body]{%
  \includegraphics[width=.25\linewidth]{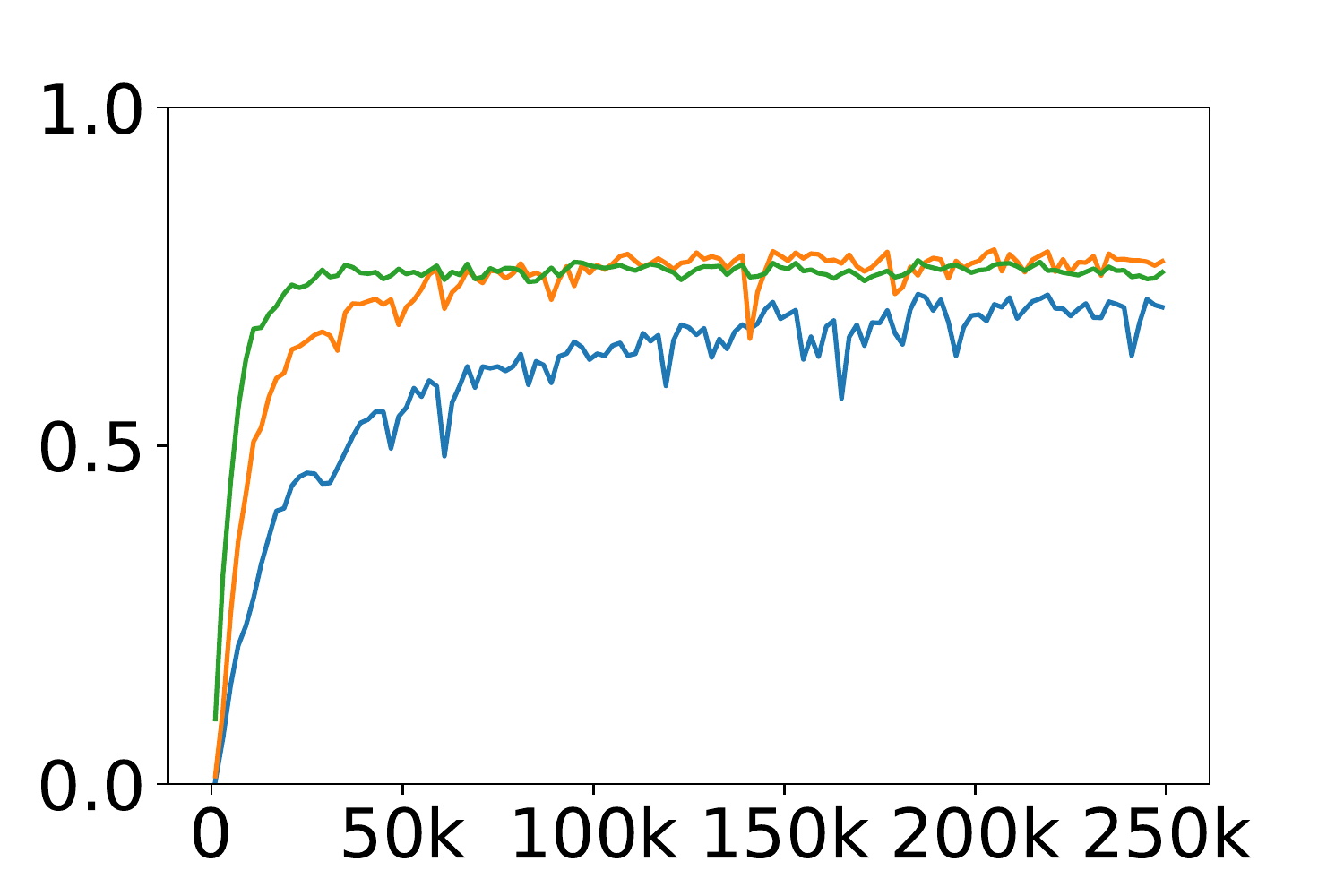}%
  \label{fig:rec_body}
}
\caption{\textbf{Precision-Recall Analysis on Head and Body Classes}: \textcolor{orange}{Orange} and \textcolor{green}{green} curves denote FixMatch with threshold 0.95 and 0.6 respectively. EnergyMatch is denoted by \textcolor{blue}{blue} curves. EnergyMatch consistently achieves higher pseudo-label precision with slightly lower recall compared with the confidence-based pseudo-labeling baselines.}
\label{fig:pr_appendix}
\end{figure}

Our ablation study, presented in Fig.~5 of the main paper, compared the precision and recall of pseudo-labels for all classes and the tail classes. In this section, we further report the results (precision and recall of pseudo-labels) for the head and body classes. 
As shown in Figure \ref{fig:pr_appendix}, pseudo-labels produced by EnergyMatch consistently achieve higher precision with slightly lower recall across head and body classes. This further justifies that the model trained with EnergyMatch is less biased towards the frequent classes compared to confidence-based pseudo-labeling.

\subsection{Robustness Experiment with Real Out-of-Distribution Samples in Unlabeled Set}
\label{app:ood}
The core idea behind our method is to pseudo-label unlabeled samples from an \textit{``in-distribution vs. out-distribution''} perspective. Thus, it would be interesting to see what happens when the unlabeled set contains real out-of-distribution examples (e.g., unlabeled images coming from a different domain). To setup this experiment, we first sample the labeled set from CIFAR10 and use the rest of CIFAR10 as well as the SVHN dataset as the unlabeled set. We construct the labeled set with 40 labeled samples in total in a balanced manner (namely, 4 instances per class)
, evaluate the trained model on CIFAR10 test data, and compare our method EnergyMatch with other SSL methods, FixMatch and FlexMatch. Selecting out-of-distribution samples (from SVHN) for pseudo labeling and training will likely decrease the model's performance on in-distribution samples (CIFAR10).

\begin{figure}[ht]
    \centering
    \includegraphics[scale=0.5]{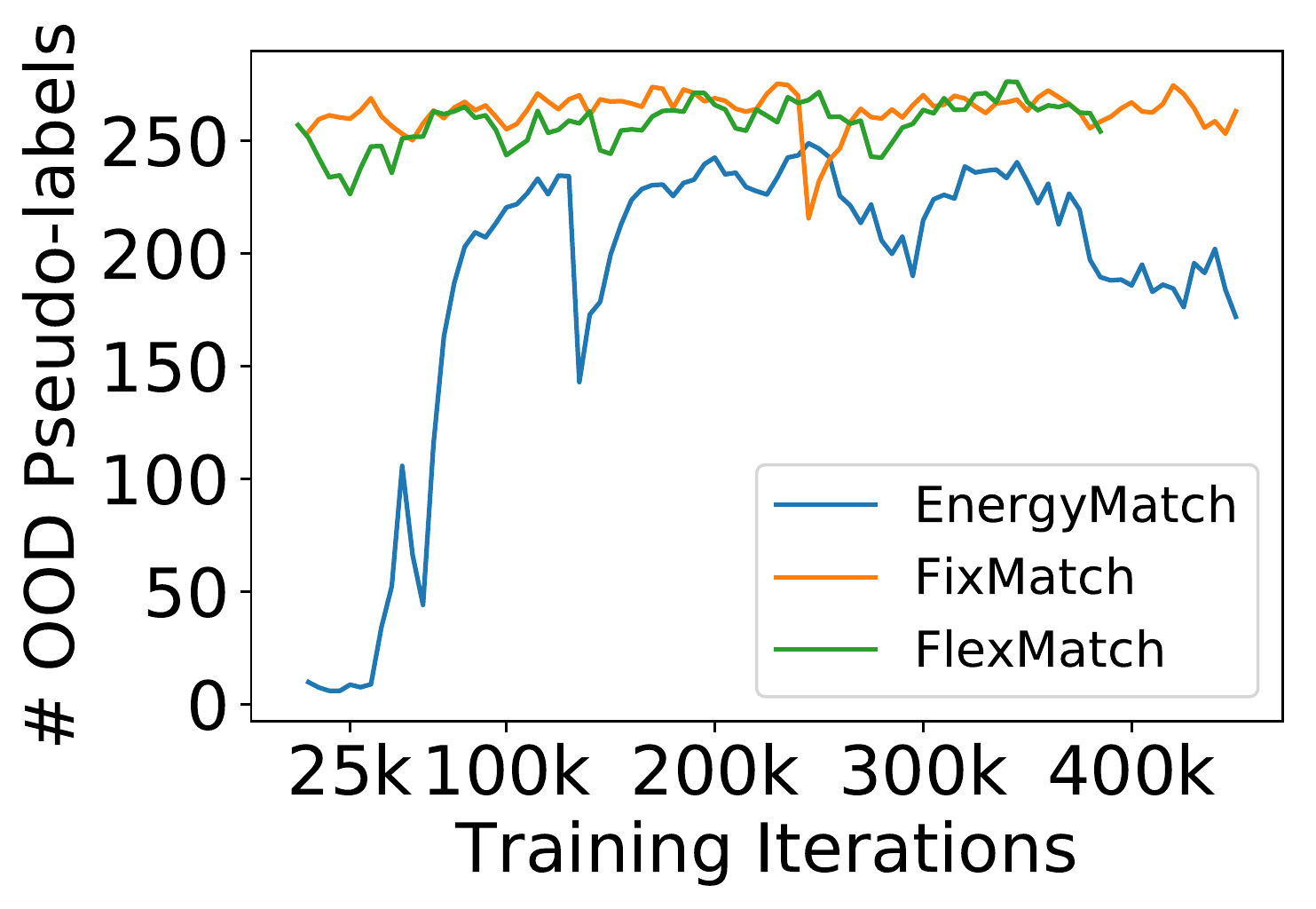}
    \caption{\textbf{Number of true OOD examples included in training:} EnergyMatch consistently includes less true OOD examples compared with FixMatch and FlexMatch.}
    \label{fig:ood}

\end{figure}

Table~\ref{tab:ood} presents the results, reported using a single random seed for all methods. The accuracy of all methods drops significantly (from 90+ to 60+) when out-of-distribution samples are presented as unlabeled data points. Even in this challenging case where the unlabeled set contains real OOD samples, EnergyMatch significantly outperforms FixMatch and FlexMatch, which demonstrates the robustness of our method. Further, as shown in Figure~\ref{fig:ood}, EnergyMatch consistently produces less pseudo-labels for OOD samples than other methods. This again shows that energy-based pseudo-labeling is more robust against OOD examples in the unlabeled set.
\begin{table}[t]
    \centering
    \caption{\textbf{Results when OOD examples appear in the unlabeled set}. We use 40 images from CIFAR10 as the labeled set and the rest of CIFAR10 and SVHN as unlabeled set. We evaluate the model on CIFAR10 test set and report the top-1 accuracy. We use energy threshold -10 for EnergyMatch. Results are reported using the same single random seed for all methods. 
    }
    \begin{tabular}{lcccc}
    \toprule
   & Labeled Set & Unlabeled Set  & Accuracy \\ 
   \cmidrule(l{3pt}r{3pt}){1-1} \cmidrule(l{3pt}r{3pt}){2-3}  \cmidrule(l{3pt}r{3pt}){4-4}
  FixMatch~\cite{sohn2020fixmatch} & \multirow{ 3}{*}{CIFAR10} & \multirow{ 3}{*}{CIFAR10} & 92.90 \\
  FlexMatch~\cite{zhang2021flexmatch} & & & \textbf{95.07} \\
  EnergyMatch & & & 94.86 \\
  \midrule
  FixMatch~\cite{sohn2020fixmatch} & \multirow{ 3}{*}{CIFAR10} & \multirow{ 3}{*}{CIFAR10 + SVHN} & 62.60 \\
  FlexMatch~\cite{zhang2021flexmatch} & & & 60.03 \\
  EnergyMatch & & & \textbf{67.02} \\
  
  \hline
  \end{tabular}
    
    \label{tab:ood}
\end{table}
\begin{table}
  \begin{tabular}{lcc}
    \toprule
        & CIFAR10 $N=40$ & SVHN $N=40$ \\ 
        \cmidrule(l{3pt}r{3pt}){2-3} 
    FixMatch~\cite{sohn2020fixmatch}     & 92.53{\scriptsize $\pm$0.28} & 96.19{\scriptsize $\pm$1.18} \\
    FixMatch-UPS~\cite{rizve2021defense} & 93.74{\scriptsize $\pm$0.87} & 97.20{\scriptsize $\pm$0.17} \\
    FixMatch-ABC~\cite{lee2021abc} & 93.16{\scriptsize $\pm$0.29} & 95.24{\scriptsize $\pm$1.64} \\
    EnergyMatch (ours) &  \textbf{94.58}{\scriptsize $\pm$0.43} & \textbf{97.78}{\scriptsize $\pm$0.05} \\
        \bottomrule
    \end{tabular}
      \caption{Comparison to FixMatch-UPS and FixMatch-ABC on CIFAR10 and SVHN with 40 labeled data in total (balanced setting).}%
  \label{tab:baseline_app}

\end{table}

\subsection{More Baseline Results (UPS and ABC)}
\label{app:baseline}
Finally, we include FixMatch-UPS and FixMatch-ABC on CIFAR10 and SVHN both with 40 labeled samples in the balanced SSL setting. The results are shown in Table \ref{tab:baseline_app}. In Table 3 of the main paper, we had included the results of UPS from the original paper, which uses a different network architecture making the results not directly comparable with other methods. In this section, we implement UPS in the FixMatch framework, providing a more direct comparison.  Specifically, this baseline produces pseudo-labels based on both the confidence score and uncertainty measurement (standard deviation over multiple Monte Carlo Dropout outputs). Our EnergyMatch maintain a noticeable edge over FixMatch-UPS, despite FixMatch-UPS requiring 10 forward passes at each iteration to compute the uncertainty metric. 


For FixMatch-ABC, we train the model with 7x larger unlabeled batch following FixMatch because using a smaller batch size as in its original setting makes the model converge in an extremely slow rate. Conceptually, in the balanced setting, FixMatc-ABC boils down to vanilla FixMatch and the empirical results validate this point; the difference between FixMatch-ABC and FixMatch is negligible. FixMatch-ABC achieves slightly better performance on balanced CIFAR10 because it generates two different augmented views for each unlabeled sample, which implicitly increases the batch size. Even so, its performance still cannot match EnergyMatch in this balanced setting. On SVHN, since the labeled set is balanced, ABC cannot leverage class distribution of labeled data to perform balanced sampling for unlabeled data even though the unlabeled data is slightly imbalanced. The performance of ABC on SVHN is even worse than FixMatch.

Taking both balanced and imbalanced settings together into consideration, our method EnergyMatch achieves the best performance among all these methods under the framework of FixMatch.

\end{document}